\documentclass[10pt,journal,compsoc]{IEEEtran}
\pdfoutput=1
\usepackage{cite}
\usepackage{url}
\usepackage{ragged2e}
\usepackage{epsfig}
\usepackage{graphicx}
\usepackage{amsmath,amssymb} 
\usepackage{subfigure}

\usepackage[linesnumbered, ruled, vlined]{algorithm2e}
\usepackage{algpseudocode}
\usepackage{graphics}
\usepackage{threeparttable}
\usepackage{color}
\usepackage[normalem]{ulem}
\usepackage{multirow}
\usepackage{float}
\usepackage{amsfonts}
\usepackage{bm}
\usepackage{array}
\usepackage[table]{xcolor}
\usepackage{colortbl}
\usepackage{pifont}
\usepackage{diagbox}
\usepackage{rotating}
\usepackage{booktabs}
\usepackage{overpic}
\usepackage{textcomp}
\usepackage{contour}
\usepackage{enumitem}
\usepackage[colorlinks=true]{hyperref}

\usepackage{multirow}
\usepackage{booktabs}%
\usepackage{algpseudocode}%
\usepackage{listings}%
\usepackage{bbding}
\usepackage{stfloats}

\RequirePackage{silence}
\definecolor{bblue}{rgb}{0,150,230}
\definecolor{mygray}{gray}{.92}

\newcommand{\supp}[1]{\textcolor{magenta}{#1}}

\graphicspath{{./Imgs/}}
\DeclareGraphicsExtensions{.jpg,.pdf,.png}
\begin{document}

\title{AUG: A New Dataset and An Efficient Model for Aerial Image Urban Scene Graph Generation}

\author{Yansheng Li,
        Kun Li,
        Yongjun Zhang,
        Linlin Wang,
        Dingwen Zhang\\
\IEEEcompsocitemizethanks{
\IEEEcompsocthanksitem Y.~Li, Y.~Zhang and L.~Wang are with the School of Remote Sensing and Information Engineering, Hubei Luojia Laboratory, Wuhan University, Wuhan, China.
(Email: yansheng.li@whu.edu.cn, zhangyj@whu.edu.cn, wangll@whu.edu.cn)
\IEEEcompsocthanksitem K.~Li is with the School of Computer Science, Wuhan University, Wuhan, China.
  (E-mail: li\_\_kun@whu.edu.cn)
\IEEEcompsocthanksitem D.~Zhang is with the Brain and Artificial Intelligence Laboratory, School of Automation, Northwestern Polytechnical University, Xi'an 710072, China.
  (E-mail: zhangdingwen2006yyy@gmail.com)

\IEEEcompsocthanksitem Corresponding author: K.~Li.}
}


\IEEEtitleabstractindextext{%
\begin{abstract}
  \justifying
  Scene graph generation (SGG) aims to understand the visual objects and their semantic relationships from one given image. Until now, lots of SGG datasets with the eyelevel view are released but the SGG dataset with the overhead view is scarcely studied. By contrast to the object occlusion problem in the eyelevel view, which impedes the SGG, the overhead view provides a new perspective that helps to promote the SGG by providing a clear perception of the spatial relationships of objects in the ground scene. To fill in the gap of the overhead view dataset, this paper constructs and releases an aerial image urban scene graph generation  (AUG) dataset. Images from the AUG dataset are captured with the low-attitude overhead view. In the AUG dataset, 25,594 objects, 16,970 relationships, and 27,175 attributes are manually annotated. To avoid the local context being overwhelmed in the complex aerial urban scene, this paper proposes one new locality-preserving graph convolutional network (LPG). Different from the traditional graph convolutional network, which has the natural advantage of capturing the global context for SGG, the convolutional layer in the LPG  integrates the non-destructive initial features of the objects with dynamically updated neighborhood information to preserve the local context under the premise of mining the global context. To address the problem that there exists an extra-large number of potential object relationship pairs but only a small part of them is meaningful in AUG, we propose the adaptive bounding box scaling factor for potential relationship detection (ABS-PRD) to intelligently prune the meaningless relationship pairs. Extensive experiments on the AUG dataset show that our LPG can significantly outperform the state-of-the-art methods and the effectiveness of the proposed locality-preserving strategy. The constructed AUG dataset will be made publicly available along with the paper \supp{\href{https://gitee.com/xiaoyibang/lpg-sgg}{https://gitee.com/xiaoyibang/lpg-sgg}}.
\end{abstract}


}

\maketitle

\IEEEdisplaynontitleabstractindextext

\IEEEpeerreviewmaketitle

\IEEEraisesectionheading{\section{Introduction}\label{sec:introduction}}
\IEEEPARstart{T}{he} scene graph is a graph-structured representation of a scene that can express the objects, the attributes of the objects, and the relationships between object pairs in the scene. According to \cite{first}, scene graphs are structured as directed graphs that are leveraged in high-level visual comprehension tasks to represent graphic relationships. As a powerful tool for high-level understanding and inferential analysis of scenes \cite{VQA}, scene graphs have attracted the attention of a growing number of researchers \cite{survey,add2,add3}. In literature, the majority of scene graph generation (SGG) datasets have been released with the eyelevel view, however, the perception of spatial relationships in ground scenes through the SGG with the eyelevel view has been severely limited by occlusion between objects. This limitation makes it challenging to interpret object positions and relationships, which easily leads to a misunderstanding of the ground scene.

\begin{figure}[t!]
\centering

    \includegraphics[width=0.98\linewidth]{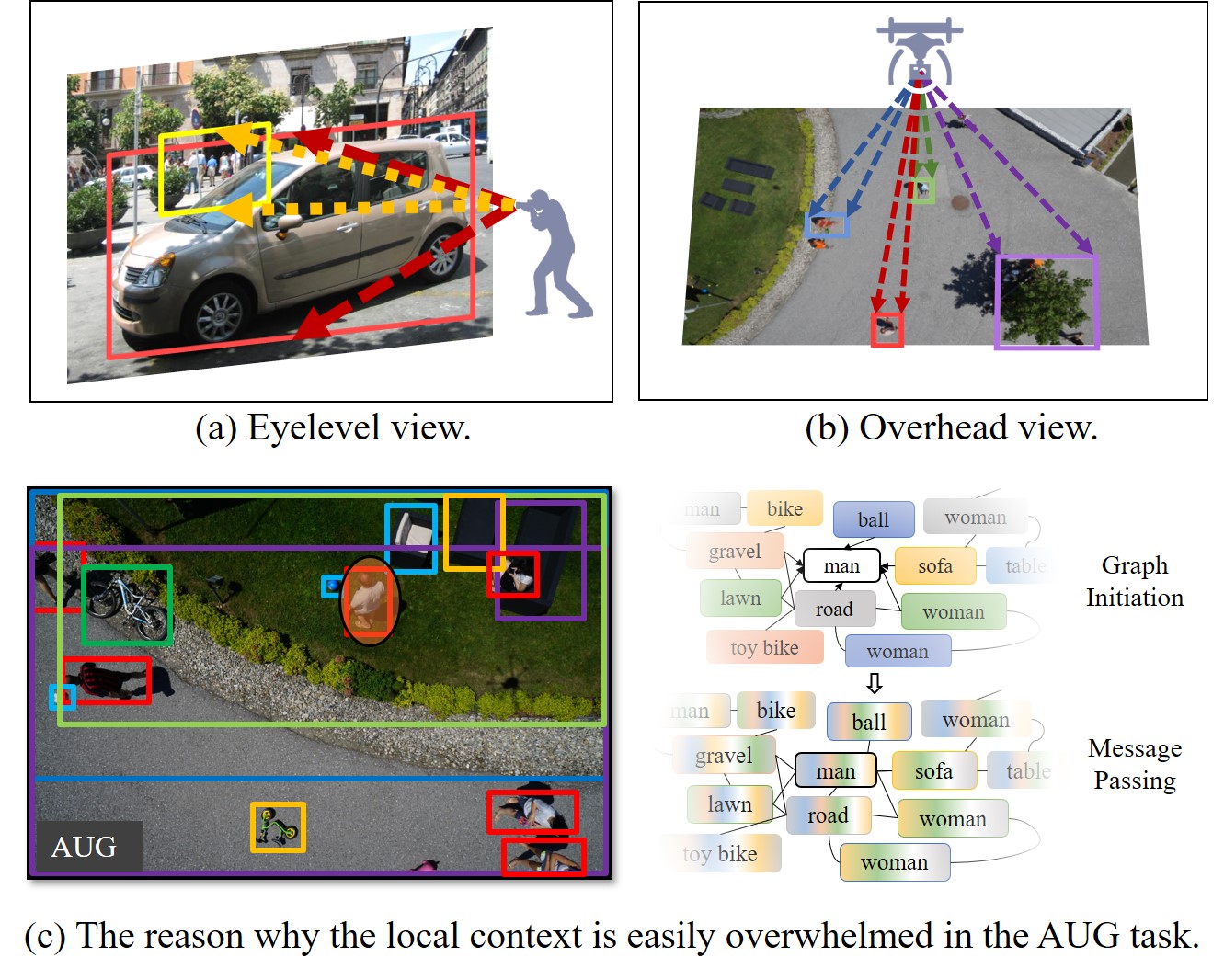}

    \caption{ The opportunities and challenges of the AUG task. (a) denotes one image with the eyelevel view from the VG dataset. (b) denotes one image with the overhead view  from the AUG dataset. (c) shows the reason why   the local context is easily overwhelmed in the AUG task.}
    \label{fig: intro}
    \vspace{-1em}  
\end{figure}

By contrast to the SGG dataset with the eyelevel view, scene graphs extracted from the SGG dataset with the overhead view are generally immune to the occlusion effects and have numerous advantages, including  smaller occlusion angles, a broader view of scene perception, and a more realistic distribution of relationships. As Fig. \ref{fig: intro} (a) and (b) display, the bounding boxes of objects from the images with the eyelevel view intersect extremely easily, while those with the overhead view do not. The position of one object projected on the screen is not coordinated with the position of the camera and the direction of view, which easily causes other objects to be occluded. Thus, in the image with the eyelevel view, objects close to the camera occupy more space, making their bounding boxes likely to intersect with those of other objects far from the camera \cite{t3}. This overlap does not imply any inherent relationship between the objects. By contrast to the majority of objects in the image with the overhead view on the ground, equidistant from the camera, and spatially distributed, reducing the issue of occlusion. Consequently, the aerial image urban scene graph generation (AUG) task based on the overhead view has an advantage in perceiving complex scenes.  The AUG task presents a new avenue for ground surface spatial relationship cognition and brings numerous opportunities, such as public safety control and early warning, emergency disaster relief, marker tracking, and so on.



In addition to the opportunities of the AUG task, it should be noted that the progress of the AUG task is heavily affected by various challenges, including the scarcity of AUG datasets, extremely nonuniform object densities, and complicated relationships between objects. These challenges severely affect the research advancement of the AUG task \cite{ODinAerialImages}. Until now, there is not any publicly available AUG dataset.  The lack of effective benchmarks has become a major obstacle to the development of SGG in aerial images. With the above considerations, this paper constructs one new AUG dataset,
which encompasses a total of 25,594 objects, 16,970 relationships, and 27,175 attributes, with an average of 63.9 objects, 42.4 relationships, and 67.9 attributes per image. In addition, we build the first benchmark for the AUG task by comparing the strengths of 9 approaches including our proposed method.

Currently, the SGG methods with the eyelevel view are advancing rapidly. These methods could be grouped into six categories: CRF-based SGG \cite{CRF1, survey}, TransE-based SGG \cite{vtranse, MATransE, motif}, CNN-based SGG \cite{GraphRCNN, HCNet}, RNN/LSTM-based SGG \cite{vctree, TDE, IMP}, GNN-based SGG \cite{EBM, DualResGCN, GraphRCNN, DGPGNN}, and Transformer-based SGG \cite{TEMPURA, HLNet}. Earlier SGG methods primarily rely on CRF, TransE, and RNN/LSTM, such as the Motifs \cite{motif} and VCTree \cite{vctree} models, which are widely recognized for their performance and become prominent baselines for model comparisons. With the development of graph neural networks, researchers explore graph theory to enhance SGG \cite{gcnsgg, GPSNet, DualResGCN }. In recent years, Transformer-based SGG methods also gain recognition, and numerous approaches are proposed \cite{TransformerADD, Relationformer}. However, these SGG methods face significant constraints due to the following challenges when employed in the image with the overhead view:

\textit{(i) Local context is easily overwhelmed.} The local context is easily overwhelmed in complex aerial urban scenes, whereas the scenes from datasets with the eyelevel view, such as Visual Genome (VG) \cite{VG}, are relatively simple and this phenomenon is not dramatically manifested. As Fig. \ref{fig: intro} (a) and (b) display, the images with the eyelevel view mostly contain primary objects in the unified background, which allows the primary objects to infer in the same context. The images with the overhead view not only do not have the primary object, but each object is highly susceptible to interference from surrounding objects (as shown in Fig. \ref{fig: intro} (c)). Therefore, the local context is easily overwhelmed in the complex aerial urban scene.

\textit{(ii) The number of potential object relationship pairs is dramatically enlarged.} While the images with the eyelevel view usually encompass a limited number of objects and relationships, the images with the overhead view encompass a broader range of objects, resulting in a significant increase in potential relationship pairs.  Existing methods for potential relationship detection (PRD) primarily adopt the IOU-based method proposed by Motif Net \cite{motif}, which simplistically determines the relationship existence through bounding box intersection. However, this approach ignores certain meaningful relationships in which the bounding boxes of object pairs do not overlap. As a whole, the number of object pairs that the AUG task has to deal with during potential relationship detection is dramatically large.

To overcome these shortcomings mentioned above, this paper proposes a more targeted approach for the AUG task in potential relationship detection and relationship prediction. To avoid the local context being overwhelmed in the complex aerial urban scene, a \textbf{l}ocality-\textbf{p}reserving \textbf{g}raph convolutional network (\textbf{LPG}) is proposed. The LPG is designed in a pattern with an initialization information preservation strategy, where the convolutional layer in the LPG integrates the non-destructive initial features of the objects with dynamically updated neighborhood information to preserve the local context under the premise of mining the global context. To address the problem that there exists an extra-large number of potential object relationship pairs but only a small part of them is meaningful, we propose a new approach for potential relationship detection, namely \textbf{a}daptive \textbf{b}oundary \textbf{s}caling factor for \textbf{p}otential \textbf{r}elationship \textbf{d}etection (\textbf{ABS-PRD}). The ABS-PRD constructs a rule set based on object classes and spatial distribution to determine whether a relationship exists between candidate objects. Extensive experiments on the AUG dataset show that the LPG can significantly outperform the state-of-the-art methods.

The \textbf{main contributions} of this paper are summarized as follows:

\begin{itemize}

\item To clearly perceive the spatial relationships of objects in the ground scene, this paper constructs the first AUG dataset, which contains 25,594 objects and 16,970 relationships.

\item  To avoid the local context being overwhelmed in the complex aerial urban scene, we propose one new LPG for the AUG task, which greatly preserves the local context of the objects. To cope with the extra-large number of relationship pairs in aerial scenes, this paper presents the ABS-PRD to effectively prune the meaningless pairs as well as keep the good ones.

\item This paper builds the first benchmark for the AUG task including our LPG and 8 state-of-the-art methods. It is noted that our LPG outperforms the existing methods.





\end{itemize}

The remainder of this paper is organized as follows. Section \ref{sec2} provides a review of related work. In Section \ref{sec3}, we introduce the AUG dataset. Section \ref{sec4} describes the proposed method of the LPG. Section \ref{sec5} presents the results of the experiments and provides a discussion. Finally, in Section \ref{sec6}, we present the conclusions of this paper.

\section{Related work}\label{sec2}

This section provides a concise review of scene graph generation methods and research on scene graph datasets in the fields of computer vision and remote sensing.

\subsection{Scene Graph Datasets}
\label{sec2.1}

To support the research of SGG methods, various types of scene graph datasets, such as image  \cite{first,openImage, VisualPhrase}, video  \cite{video}, 3D  \cite{3DDataset}, etc, have been released, however, the AUG datasets are very scarce. Scene graph datasets with the  eyelevel view included RW-SGD \cite{first}, Visual Phrase \cite{VisualPhrase}, VG \cite{VG}, its optimized versions VG150 \cite{openImage} and VrR-VG (Visually-Relevant Relationships Dataset) \cite{VrRVG}. Moreover, CAD120 \cite{CAD120} and Action Genome \cite{ActionGenome} were video datasets, whereas 3DSSG \cite{3DSSG} and \cite{3DDataset} were examples of 3D datasets.

Among them, VG150 and VrR-VG were built on VG; VG150 eliminated objects with poor quality, overlapping bounding boxes, and unclear object names in the annotation, while VrR-VG highlighted visually relevant relationships and suppressed the long-tail effect of relationships. CAD120 \cite{CAD120} and Action Genome \cite{ActionGenome} were video-based datasets, which can be used for analyzing spatiotemporal scene graph tasks. In the 3D scene graphs category, there were 3DSSG \cite{3DSSG}, Scene HGN \cite{t6}, and 3DDataset \cite{3DDataset}. These datasets were useful for various 3D scene understanding tasks, such as robot navigation, 3D indoor scenes, and augmented and virtual reality applications. GRTRD \cite{GRTRD} was a high-resolution remote sensing image intended to identify geo-objects such as building areas, roads, meadows, hardened surfaces, bare land, river, water, and railway, among others. The dataset comprised 3,200 images with a resolution of 0.5m and a size of 600 $\times$ 600 and includes a total of 19,904 annotated geo-objects and 18,602 geospatial relations; 12 kinds of geo-objects and 19,904 object instances and 18,602 relationship instances were available. RSSGD \cite{RSSGD} was used to enhance the recognition capability of remote sensing objects by annotating attributes such as shape and color based on object and relationship annotation, using the RS image caption dataset, which comprised 10,921 images with a size of 224 $\times$ 224.



By contrast to images with eyelevel view and satellite imagery, photography platforms taken by UAV integrated two advantages with high spatial resolution and wide field of view, benefiting the recognition of smaller size objects and inferring semantic relationships. In particular, the object annotation in the AUG dataset included humans, so its relationship annotation contained geospatial relationships and human-human interaction/human-object interaction  \cite{hhi1,hhi2}. Although GRTRD \cite{GRTRD} and RSSGD \cite{RSSGD} have high resolution, as high-resolution remote sensing satellite images, they can only understand and infer the spatial relationship of large-scale objects. This was insufficient for small object recognition, which included people, vehicles, plants, balls, windows, etc. The AUG task is in urgent need of development, but there are currently no publicly available AUG datasets.


\subsection{Scene Graph Generation Methods}
\label{sec2.2}

The SGG task requires understanding the visual objects and their semantic relationships from one given image, including the positions of objects, the categories of objects, and the categories of relationships between objects.

The concept of scene graph was first introduced by Johnson et al. \cite{first}, along with the Real-World Scene Graphs (RW-SGD) dataset. Since then, several approaches have been developed in the field of scene graph generation, such as Motif Net proposed by Zellers et al. \cite{motif}, which used Bidirectional LSTMs to propagate global contextual information and identified regularly appearing substructures in scene graphs called motifs. The VCTree \cite{vctree} constructed a dynamic tree from visual features of objects, while Tang et al. \cite{TDE} proposed an inference algorithm based on Total Direct Effect (TDE) for unbiased visual reasoning tasks. To address the long-tail phenomenon, Chiou et al. \cite{DLFE} proposed Dynamic Label Frequency Estimation (DLFE), which estimated label frequencies and obtained unbiased probability distributions of categories. GPSNet \cite{GPSNet} augmented the node features with node-specific contextual information and encoded the directionality of edges; while Suhail et al. proposed an energy-based learning framework (EBM) \cite{EBM} that performs structure-aware learning and optimizes complete scene graphs. IETrans \cite{IETrans} tackled data distribution problems, including long-tail distribution, and semantic ambiguity issues through data-driven approaches. To mitigate bias and generate unbiased scene graphs, existing methods focused primarily on two aspects: the training and inference stages. Adaptive weighting \cite{adaptiveweighting1, adaptiveweighting2}, adaptive message propagation mechanisms \cite{message1, message2}, resistance training \cite{resistancetraining}, and language-guided strategies \cite{languagestrategies} are some of the techniques employed. Several methods based on prior knowledge (such as language prior \cite{lp1, lp2}, statistical prior \cite{sp1, sp2}, and knowledge priors \cite{kp1, kp2, kp3}) have been proposed.

In recent years, many approaches for scene graph generation have focused on two directions, namely internal information mining and external information exploitation. For instance, CoRF, a composite relational field (CoRF) representation-based method \cite{CoRF}, was proposed by George Adaimi et al. that allowed relationship detection to be converted into an intensive regression and classification task. Xianjing Han et al. \cite{pp} developed an offline pattern-predicate correlation mining algorithm that adopted the divide-and-conquer approach to discover similar predicates sharing the same object interaction pattern. Furthermore, approaches like SSDN \cite{SSDN} and Khan et al. \cite{changshi2} included external information on relational semantics or common-sense knowledge about semantic elements to supplement the learning. As memory consumption and inference speed were also important concerns, Yuren Cong et al. proposed Relation Transformer (RelTR) \cite{RelTRRT}, an end-to-end sparse scene graph generation model that utilized only visual appearance to predict scene graphs and improved inference speed by 60\% without significant loss in accuracy. In summary, Motif Net \cite{motif}, VCTree \cite{vctree}, GPSNet \cite{GPSNet}, and other models \cite{MATransE, DualResGCN, CRF1} were the backbone models, mainly used for feature extraction, while TDE \cite{TDE}, DLFE \cite{DLFE}, and other models \cite{IETrans, PUM} were optimized strategies for the prediction of tasks based on other backbone networks.

Numerous scene graph generation methods have been proposed, but they cannot be applied directly to remote sensing. Most methods employed Motif Net's strategy for potential relationship detection, which used intersection over union (IOU) to determine whether two objects have a potential relationship. However, remote sensing images often had a large and dense number of objects, and the spatial location relationship between them was complex. Consequently, the IOU-based potential relationship detection method was not highly applicable to remote sensing images.

To date, no public methods existed in the field of AUG, and only two known methods could generate natural language descriptions of remote sensing images (including prediction of relationship predicates between objects). To address the model blindness when searching the semantic space, RSSGG\_CS\cite{RSSGGCSRS} used statistical knowledge of relational predicates and Wikipedia texts as supervision during the training phase. SRSG\cite{SRSG}, based on remote sensing image segmentation, employed a multi-branch structure that embedded morphological features of object pairs and mapped them to the semantic space for relationship prediction.

\section{Aerial Image Urban Scene Graph Dataset}\label{sec3}

\begin{figure*}[ht]
    \centering
    \includegraphics[width=0.99\textwidth]{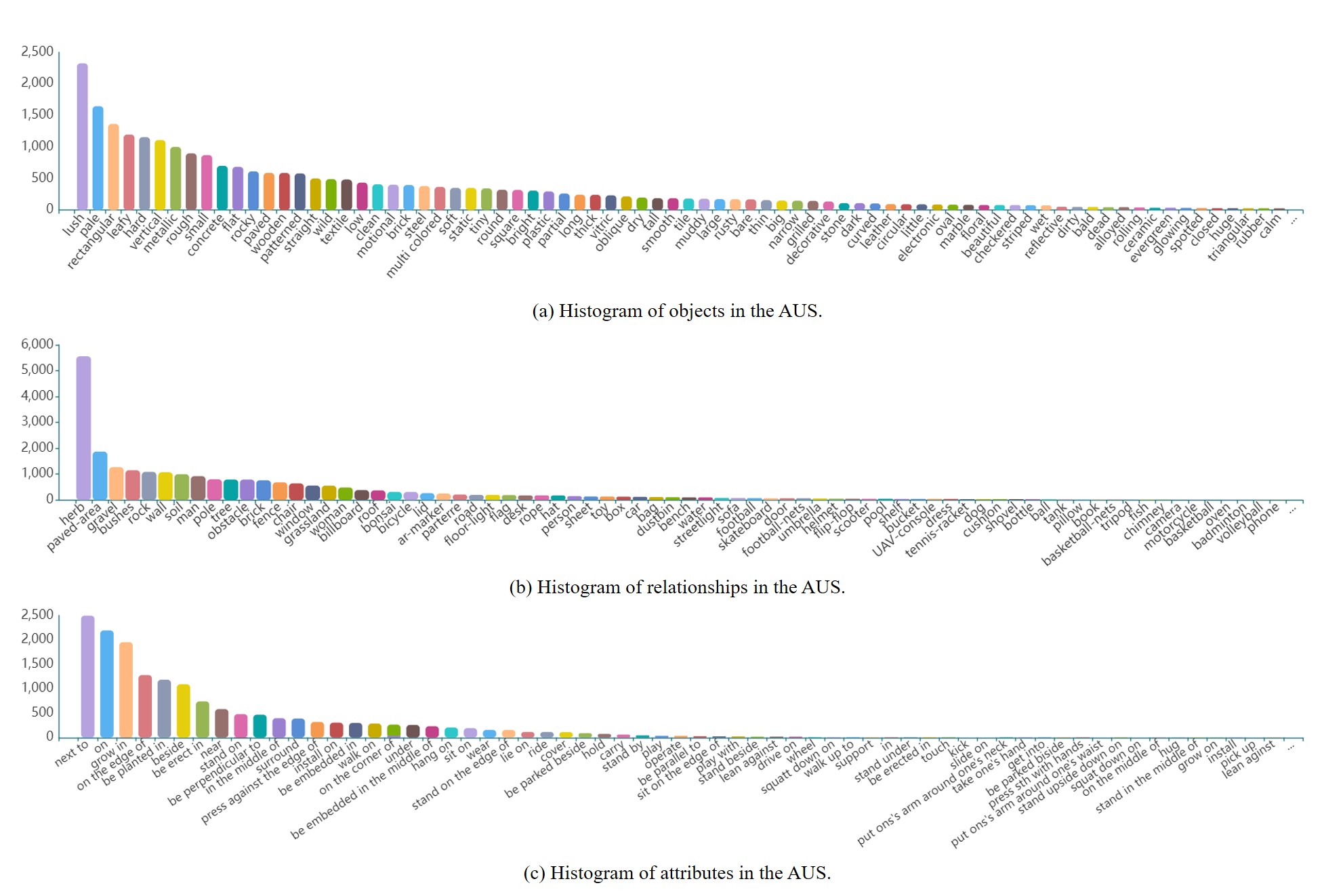}
    
    
\DeclareGraphicsExtensions.
\caption{Quantity distribution of objects, relationships,  and attributes on AUG. (a) is the quantity distribution of objects. (b) is the quantity distribution of relationships. (c) is the quantity distribution of attributes.}
\label{fig:histogram}
\end{figure*}

The AUG dataset is constructed based on the published semantic drone dataset (SDD) \cite{SDD}. SDD is an aerial image urban dataset captured by UAVs, providing an overhead view from 5 to 35 meters. In total, the AUG dataset contains 400 images, and each image contains 6,000 $\times$ 4,000 pixels (24Mpx). Object information is annotated through horizontal bounding boxes, attribute category annotations, and relationship category annotations for object pairs, with an average of 63.9 objects, 42.4 relationships, and 67.9 attributes per image. Each object has on average 2.45 attributes. The AUG dataset includes both geospatial relationships and human-human interaction/human-object interaction \cite{hhi1}, with 1,562 human objects and 2,147 human-human interaction/human-object interaction relationships. Overall, 25,594 objects, 16,970 relationships, and 27,175 attributes are manually annotated in the AUG dataset, which is capable of training and validating AUG methods besides SGG methods.

\subsection{Data Collection and Labeling}
\label{sec3.1}

\begin{figure*}[ht]
  \centering
  \includegraphics[width=0.95\textwidth]{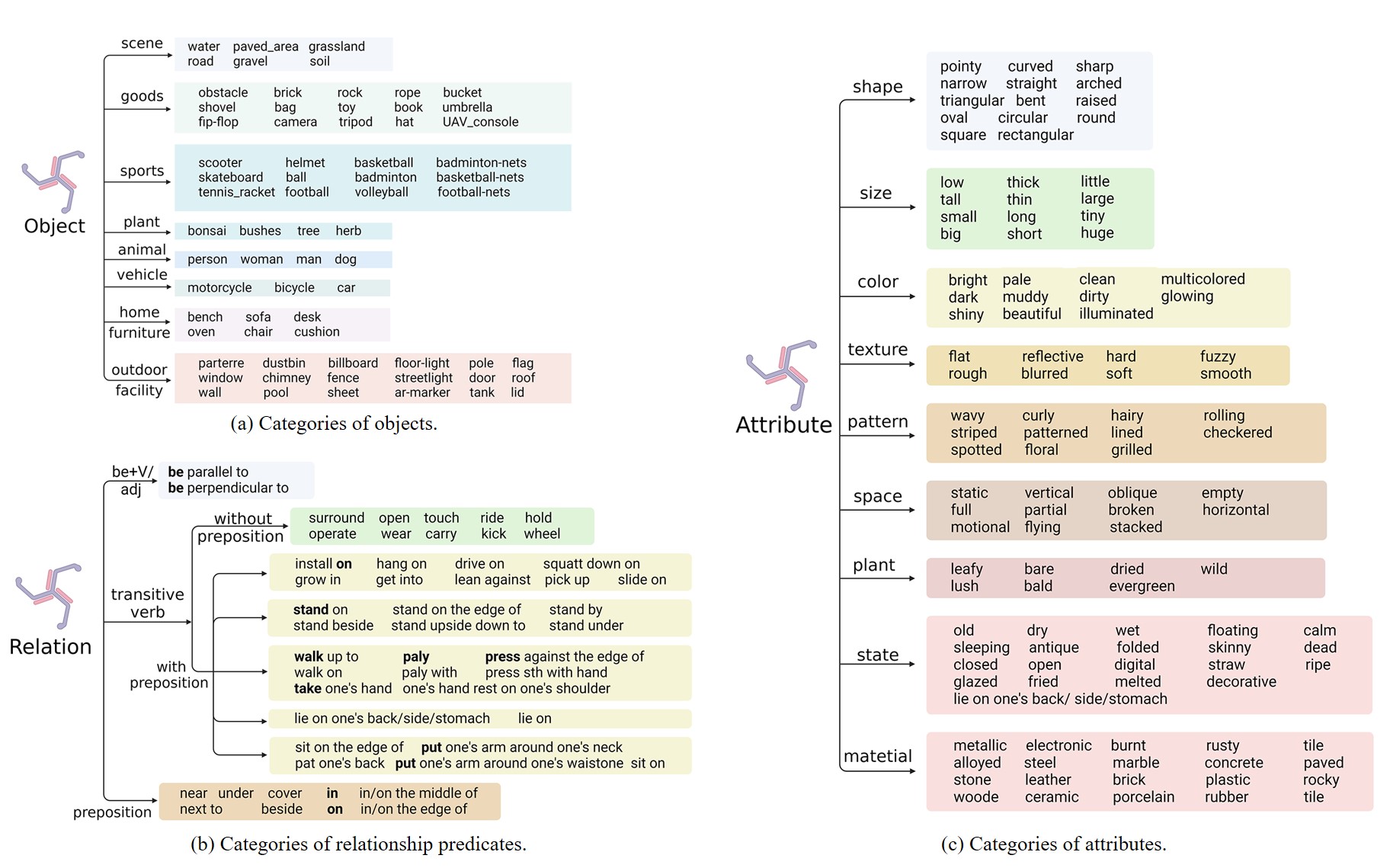}
  \caption{ Annotation categories for objects, relationships, and attributes on the AUG dataset. (a) shows the categories of objects, we divide the objects into area-wide classes and independent individual objects, and the individual objects are divided according to their own properties. (b) shows the categories of relationship predicates according to grammatical rules in the category of relations, which allows adding more regular relational predicates according to the rules. (c) The categories of attributes are divided into nine aspects: shape, size, color, texture, texture, space, plant description, state, and material.}
   \label{fig: categories}
  
\end{figure*}

The frequency of object categories in the AUG dataset is calculated and set, as shown in Fig. \ref{fig:histogram} (a) and \ref{fig: categories} (a). During the analysis of the full dataset, the 14 object categories from SDD are retained, and polygon labeling is slightly modified by dividing some generalized categories into finer categories. For example, "vegetation" is subdivided into four "plant" types to enable the exploration of the impact of different vegetation distributions on the scene graph generation task. The category "person" has been split into "woman" and "man" due to the absence of gender information, and "person" has been added to refer to situations in which gender cannot be determined. Categories such as "fence-pole" and "dirt" have been removed due to their small size, limited practical significance for detection, and high labeling cost. Additionally, the original category "obstacle" has been divided into numerous unlabeled objects, and categories have been reset based on actual conditions. The introduction of these categories significantly improves the semantic information of the scene graph generation task.


\begin{table*}[ht!]
\centering

\renewcommand\arraystretch{1.2}
\setlength{\tabcolsep}{5pt}
\caption{Characteristics of AUG dataset compared with other public image-based SGG datasets. \#OPI counts objects per image, \#RPI counts relationships per image, \#API counts attributes per image, and DupFree checks whether duplicated object groundings are
cleaned up.}
\resizebox{\textwidth}{!}{%
\begin{tabular}{ccccccccccc}
\hline
Dataset & View type  & \#OPI & \#RPI & \#API & \# ObjCls & \#RelCls & \#AttrCls & DupFree & Multi Attr & Connected Graph \\ \hline
VRD  \cite{VRD} & Eyelevel  & - & 7.6 & - & 100 & 70 & - & \XSolidBrush & \XSolidBrush & \XSolidBrush \\
Scene Graph \cite{SceneGraph} & Eyelevel & 13.8 & 22 & - & 266 & 23K & - &\XSolidBrush  & \XSolidBrush & \XSolidBrush \\
Visual Phrase \cite{VisualPhrase} & Eyelevel  & 1.2 & 0.6 & - & 8 & 17 & - &\XSolidBrush  &  \XSolidBrush & \XSolidBrush \\
HCVRD \cite{HCVRD} & Eyelevel  & - & 4.9 & - & 1.8K & 927 & - & \XSolidBrush &  \XSolidBrush & \XSolidBrush \\
VG150 \cite{VG150} & Eyelevel  & 8.4 & 4.7 & - & 150 & 50 & - & \XSolidBrush &  \XSolidBrush & \XSolidBrush \\
VrR-VG \cite{VrRVG} & Eyelevel  & 4.7 & 3.4 & - & 1.6K & 117 & - & \XSolidBrush & \XSolidBrush & \XSolidBrush \\
VG  \cite{VG} & Eyelevel  & 16 & 18 & 16 & 76.3K & 47 & 15.6K & \XSolidBrush & \Checkmark & \XSolidBrush \\
RW-SGD \cite{RW-SGD} & Eyelevel & 18.6 & 22.4 & 22 & 6.7K & 1.3K & 3.7K & \XSolidBrush & \Checkmark & \XSolidBrush \\
PSG \cite{PSG} & Eyelevel  & 11 & 5.6 & - & 133 & 56 & - & \Checkmark & \XSolidBrush & \Checkmark \\ \hline
AUG (Ours) & Overhead & 63.9 & 42.4 & 67.9 & 76 & 61 & 108 & \Checkmark & \Checkmark & \Checkmark \\ \hline
\end{tabular}%
\label{tab: dataset1}}
\end{table*}

\begin{figure*}[ht]
\centering
\includegraphics[width=0.88\linewidth]{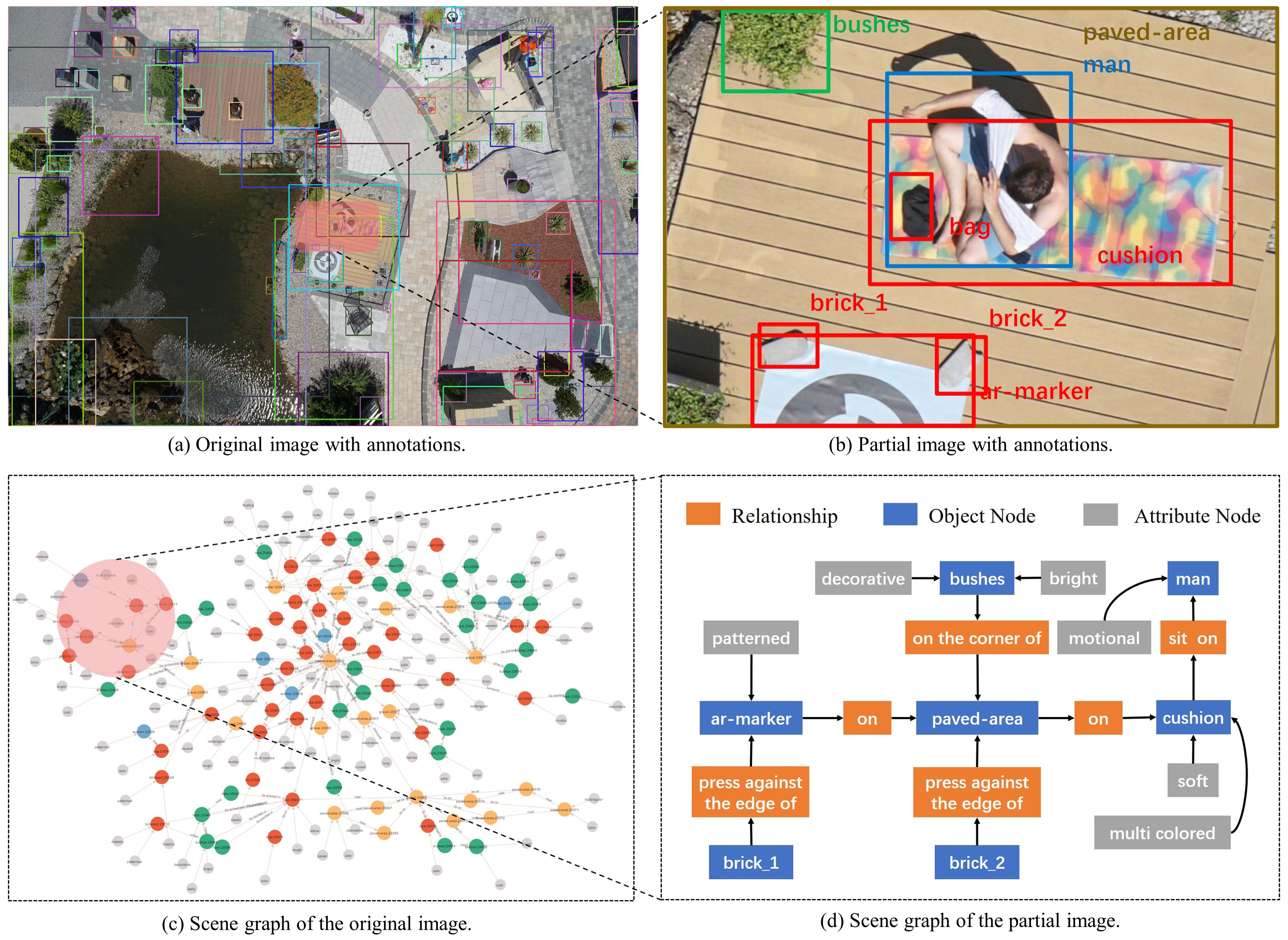}
	
\caption{
Visualization of annotations of the AUG dataset. (a) and (c) show the original image and its corresponding scene graph. (b) and (d) show the details of the red areas in (a) and (c). The colored nodes and the gray nodes indicate objects and attributes, respectively, and the directed edges between the nodes indicate that a relationship exists between the two nodes. A pointing to B indicates that A is the subject and B is the object. Specifically, an attribute node pointing to an object node indicates that this object has this attribute.
}	
\label{fig 5}
\end{figure*}

As shown in Fig. \ref{fig:histogram} (b), relational categories containing semantic meaning surpass the number of simple categories lacking semantic meaning, such as "on", "near", and "in". These words are detailed as "stand/sit/lie on" and "on the edge/corner of" by modifying prepositions and verbs to add semantic meaning. Referring to Fig. \ref{fig: categories} (b), the vocabulary for relational annotation is established based on the words' relevance and the syntactic rules of phrases. This facilitates the increase of relational categories and knowledge graph embedding after the expansion of the dataset, based on the translation invariance assumption. Using the base vocabulary as the parent node and setting relational sub-nodes with higher semantic information can improve data transplantability. As depicted in Fig. \ref{fig: categories} (b), "stand", as an intransitive verb, serves as the relational root for "stand on", "stand by", "stand under", "stand on the edge of", and other similar relationships. If new data contain the phrase "someone is standing in a certain area in the middle of", the category "stand in the middle of" can be added to describe "stand in the area" (where "in the middle of" is an existing category in this dataset). Some relational phrases use "sth" and "one's" to indicate the subject and object's contextual position in the relationship. Normalizing the representation of these relational predicates aids in acquiring labeled words through VTransE \cite{vtranse, t2} and Knowledge Graph Embedding models to obtain vector features of labeled words.

Fig. \ref{fig:histogram} (c) displays the distribution of attribute categories, while Fig. \ref{fig: categories} (c) illustrates the attribute categories within the AUG dataset, which are divided into nine aspects: shape, size, color, texture, space, plant description, state, and material. By performing tasks such as attribute annotation, object detection, relationship detection, and relationship classification, model performance is expected to be improved. It should be noted that color is not included in the attribute annotation of the AUG dataset because attributes should provide reference information for relationship detection and scene graph generation. Color, shape, and other attributes can be obtained directly from the image, for example, color can be obtained from images of the visible light type.

\subsection{Dataset Statistics and Analysis}

The AUG dataset contains a total of 400 images with a format size of 4,000 $\times$ 6,000px (24Mpx) and encompasses  25,594 objects, 16,970 relationships, and 27,175 attributes. Notably, 11,090 objects within the AUG dataset contain 27,175 attributes, with an average of 2.45 attributes per object annotation. The statistics for displaying the annotation of objects, attributes, and relationships are provided below:

Fig. \ref{fig:histogram} shows that there is a long-tail effect in the distribution of the categories of objects, relationships, and attributes. The highest frequency relationships are "next to," "on," and "grow in," and have strong associations with high-frequency objects such as "herb," "paved area," and "gravel," and high-frequency attributes such as "lush," "leafy," and "hard." The long-tail effect presents a significant challenge for visual relationship recognition. Certain object categories exhibit spatial connections, for instance, plants typically grow in soil, while grass and concrete pavements are found in similar settings. Meanwhile, recognizing the relationship categories between people and objects can be more complex and varied, including actions like "sit on," "walk on," "carry," "hold," "kick," and "press sth with hands". Spatial relationships are particularly challenging for predicting mixed human-human/human-object relations, which presents another hurdle for the relationship prediction task of the AUG dataset.

The statistical results of the common scene graph dataset for visuals are shown in Table \ref{tab: dataset1}, from which it is obvious to compare the differences of the AUG dataset in terms of the view type, image size, degree of attribute labeling, and other factors. In the scene graph, the term \textbf{Connected Graph} means that any two points (i.e., objects) are connected. In a connected graph, the flow of information between nodes is smoother and more conducive to context awareness and relationship discernment. Moreover, on average, each image in the AUG dataset contains 63.9 objects, with a maximum of 395 objects in a single image. In comparison, RW-SGD contains 18.6 objects per image, VrR-VG contains 4.7, VG150 contains 8.4, and VG contains 16 objects per image (with duplicated bounding boxes). Due to the large number of objects per image in the AUG dataset, i.e. a \textbf{high-density object} annotated dataset, a fully permutated combination approach for the initialization of the relationship graph may impair computational efficiency and model performance. Thus, developing a relationship extraction module to extract object pairs with possible relationships presents a fresh challenge for the remote sensing scene graph generation task. Furthermore, compared to the datasets with the eyelevel view, the AUG dataset offers a full spectrum of environmental information and spatial distribution for each object. They provide the potential for perceiving complex scenes and making rational decisions.


Object annotation can be applied to object detection, instance segmentation, panoramic segmentation, semantic segmentation, and other object detection tasks. Attribute annotation is the annotation of partial objects in an image, and there may be multiple attribute classes for one given object, and the attributes of an object  are annotated by the experience and knowledge of the annotator. Since attributes are shared, annotating the attributes of some objects in a class can achieve the requirement of learning attribute features and reduce the annotation time and cost. As illustrated in Fig. \ref{fig 5}, AUG provides visualizations of image annotation information. Upon establishing the relationships between the objects through annotation, attributes are added to the object nodes. This results in the formation of a scene graph data structure, comparable to the knowledge graph model. By evaluating the connection component, we can verify that all objects in the relational scene graph are related to at least one other object.

\section{Locality-preserving graph convolutional network for aerial image urban scene graph generation}
\label{sec4}

In this study, multi-label attribute annotation and object annotation  are used for the attribute prediction task and object detection task, respectively, then an efficient potential relationship detection method based on adaptive bounding box scaling factor is introduced, and finally, their application to scene graph generation task is presented. In view of the current trend of scene graph generation tasks, this paper proposes a locality-preserving graph convolutional network (LPG) for the AUG task, which fully preserves the local context and integrates the updated neighborhood features. The workflow of the LPG method is visualized in Fig. \ref{fig: model}. Firstly, the visual feature map of each object is extracted via the Resnet50, which serves as the backbone network for the Cascade R-CNN \cite{cascade}. Secondly, the feature map of the Feature Pyramid Networks (FPN) \cite{FPN} is utilized for two usages. The first usage is for attribute training and prediction, and the second usage is for obtaining the final object classes and bounding boxes via the Cascade R-CNN module. After acquiring the position, category, and attribute information of each object using the object detection model and multi-label attribute classification model, ABS-PRD filters all object pairs to obtain $k$ pairs of potential relationships. Through message passing and node updating, the locality-preserving graph convolutional network retrieves node representations with contextual information that merges with the node neighborhood information while preserving the local initial node information. Finally, the positions and categories of object pairs are predicted by combining potential relationship information, thereby generating the corresponding scene graph.

\subsection{Potential Relationship Detection Based on Adaptive Bounding Box Scaling Factor}
\label{sec4.1}

The traditional method of Potential Relationship Detection (PRD) considers object pairs with overlapping bounding boxes as potential relationship pairs (IOU-PRD). However, aerial images have the characteristics of large scale, multiple objects, and complex situations, which include the long-range interaction relationship between people. Since the IOU-PRD is unable to detect long-range object pairs with disjoint boxes of objects, the potential relationship detection method based on the adaptive bounding box scaling factor (ABS-PRD) method is proposed to solve these problems.


\begin{figure}[ht]
    \centering
    \includegraphics[width=.9\columnwidth]{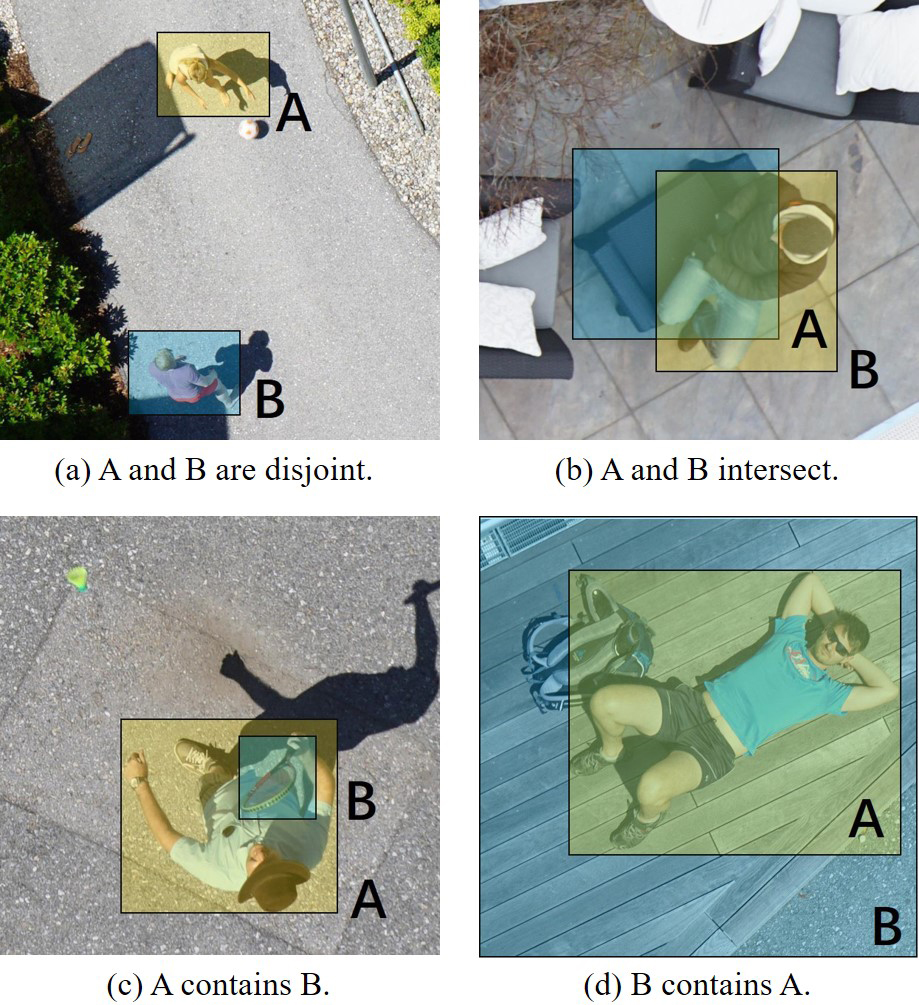}




	\caption{Four kinds of bounding boxes about ABS.}
	
	\label{fig: BBox four kinds}
\end{figure}

\begin{figure}[ht]
\centering
\includegraphics[scale=0.29]{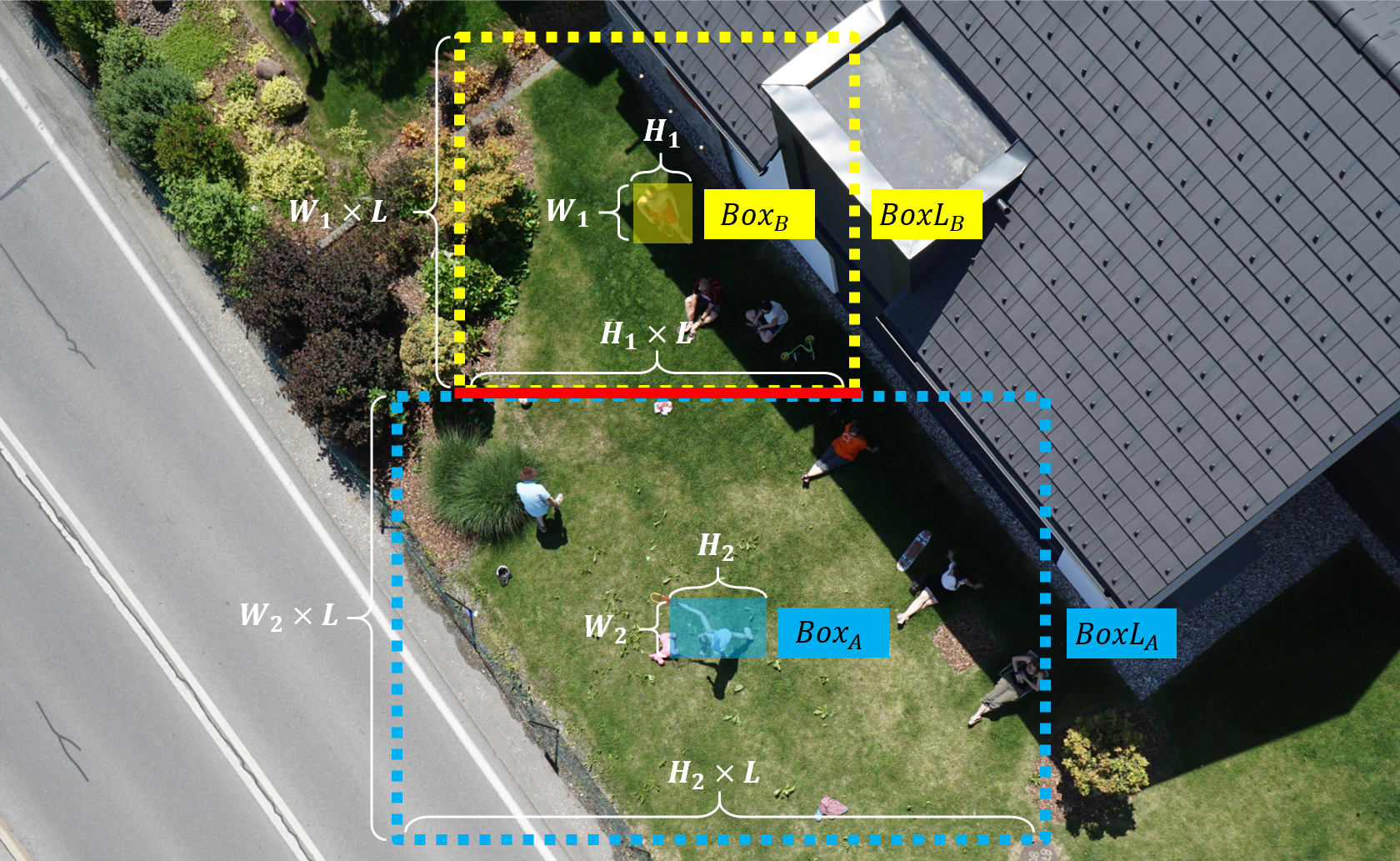}
\caption{An example that needs to be zoomed in, $Box_A$ and $Box_B$ are disjoint bounding boxes corresponding to two people in the image, and the bounding boxes are zoomed in by  the same ratio $L$ until they just intersect (red lines are used to indicate the intersection of the bounding boxes).}
\label{fig abbsT}
\end{figure}


Two objects (A and B) have four relative positions in 2D Euclidean space (refer to Fig. \ref{fig: BBox four kinds}). Object A and object B can be either non-intersecting (as in Fig. \ref{fig: BBox four kinds} a), intersecting but not containing the other object (as in  Fig. \ref{fig: BBox four kinds} b), object A containing object B (as in Fig. \ref{fig: BBox four kinds} c), or object B containing A \ref{fig: BBox four kinds} d). The study then records the position-related statistics of the category pair $<Cate_A, Cate_B>$ of the dataset dynamically. The symbols $Cate_A, Cate_B$ mean the category of object A and object B respectively, while $Box_A, Box_B$ represent the bounding boxes of A and B, respectively. Finally, a dictionary $M$ of category pairs $<Cate_A, Cate_B>$  is created, where it contains four keys that are $M_C(A, B)$, $M_C(B, A)$, $M_d(A, B)$, $M_u(A, B)$. Their specific meanings are outlined in Table \ref{tab:M}.

\begin{table*}[ht]
\centering
\small  
\renewcommand\arraystretch{1.3}
\setlength{\tabcolsep}{15pt}
\caption{Instances of the dictionary $M$ corresponding to two object categories $A$ and $B$ given, $A, B$ are two categories of objects.}
\resizebox{\textwidth}{!}{%
\begin{tabular}{cccc}

\hline
Key & Definition                                                                      & Data Type & Example  \\ \hline
$M_C(A,B)$   & \begin{tabular}[c]{@{}c@{}}The number of times the bounding box of\\
$Cate_A$ surrounds the bounding box of $Cate_B$\end{tabular}      & Int                & 231               \\
$M_C(B,A)$   & \begin{tabular}[c]{@{}c@{}}The number of times the bounding box of\\
$Cate_B$ surrounds the bounding box of $Cate_A$\end{tabular}      & Int                & 43                \\
$M_d(A,B)$   & \begin{tabular}[c]{@{}c@{}}The \textbf{minimum} zoom-out factors $M_d(A,B)$ of\\
the intersecting bounding boxes $Box_A,Box_B$\end{tabular}     & List               & 0.118, 0.012, 0.445, ... \\
$M_u(A,B)$   & \begin{tabular}[c]{@{}c@{}}The \textbf{maximum} zoom-in factors $M_u(A,B)$ of \\
the non-intersecting bounding boxes $Box_A,Box_B$\end{tabular}  & List               & 1.416, 1.423, 2.428, ...    \\ \hline
\end{tabular}%
\label{tab:M}}
\end{table*}

\begin{table*}[htbp]     
\caption{Four typical examples are given in the table with the statistical results of the ABS-PRD algorithm. '$S$' means a subject,'$O$' means an object, $M_C(SO)$ means "the times of subject contains object", $M_C(OS)$ means "the times of object contains subject", the relation pair "A-B" means A is the subject and B is the object.}  
\renewcommand\arraystretch{1.2}
\setlength{\tabcolsep}{15pt}
\resizebox{\textwidth}{!}{%
\begin{tabular}{ccccc}
    \hline
    
    Relation Pair  & $M_C(S,O)$   & $M_C(O,S)$ & Distribution of $M_d(S,O)$ & Distribution of $M_u(S,O)$ \\ 
    
    \hline
    
    \multirow{2}{*}
    {woman-man}
    &
    \multirow{2}{*} 
    {0\%}
    &
    \multirow{2}{*} 
    {0\%}
    & 
    \begin{minipage}[b]{0.35\columnwidth}
        \raisebox{-.5\height}
        {
            \includegraphics[width=\linewidth]         {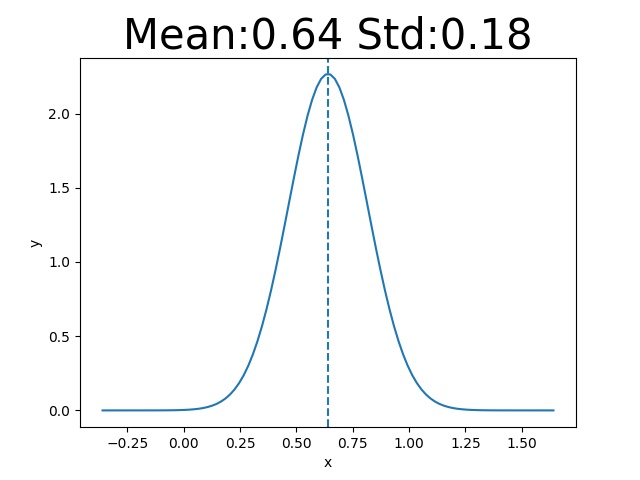}
         }
    \end{minipage}    
	& 
    \begin{minipage}[b]{0.35\columnwidth}
		\raisebox{-.5\height}
		{\includegraphics[width=\linewidth]{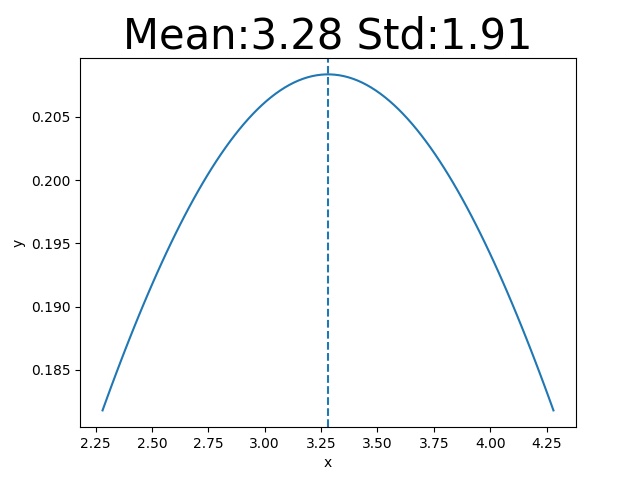}}
    \end{minipage}    
        
    \\  
    
    \hline
    \multirow{2}{*}
    {paved\_area-gravel}
    & 
    \multirow{2}{*}
    {36\%}
    & 
    \multirow{2}{*}
    {64\%}
    & 
    \begin{minipage}[b]{0.35\columnwidth}
    	\raisebox{-.5\height}
    	{\includegraphics[width=\linewidth]{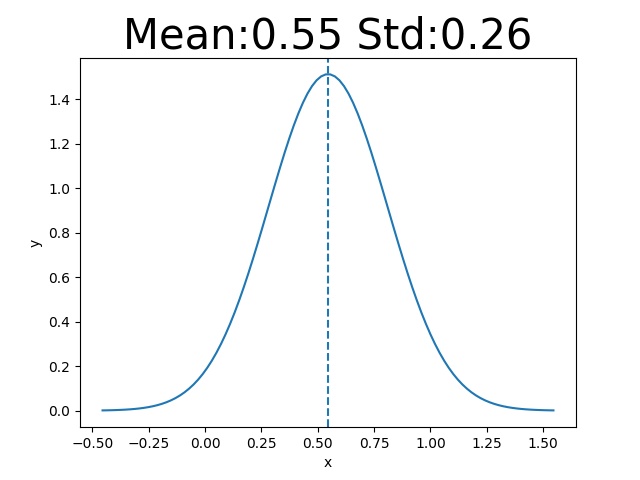}}
	\end{minipage}    
	& 
	\begin{minipage}[b]{0.35\columnwidth}
		\raisebox{-.5\height}
		{\includegraphics[width=\linewidth]{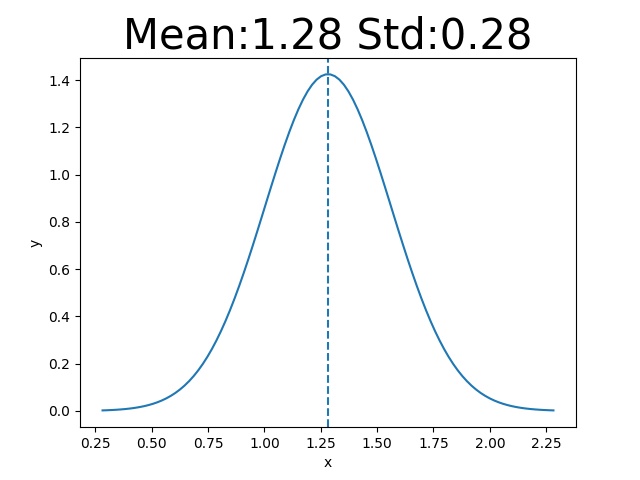}}
    \end{minipage}    
        
    \\  
    
    \hline
    \multirow{2}{*}
    {man-hat}
    & 
    \multirow{2}{*}
    {100\%}
    & 
    \multirow{2}{*}
    {0\%}
    & 
    \begin{minipage}[b]{0.35\columnwidth}
    	\raisebox{-.5\height}
    	{\includegraphics[width=\linewidth]{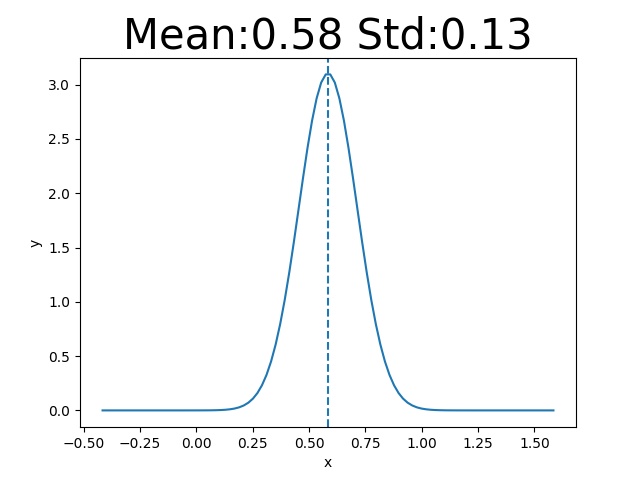}}
	\end{minipage}    
    & 
    \multirow{2}{*}
    {None}
    
    \\  
    
    \hline
    \multirow{2}{*}
    {man-grassland}
    & 
    \multirow{2}{*}
    {0\%}
    & 
    \multirow{2}{*}
    {100\%}
    & 
    \begin{minipage}[b]{0.35\columnwidth}
    	\raisebox{-.5\height}
    	{\includegraphics[width=\linewidth]{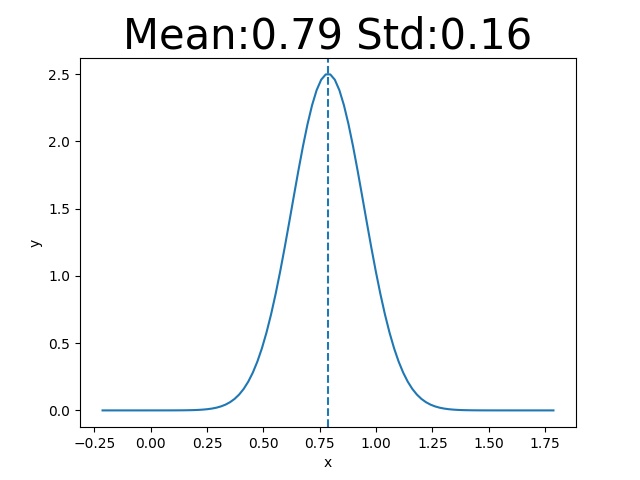}}
	\end{minipage}   
	& 
	\begin{minipage}[b]{0.35\columnwidth}
		\raisebox{-.5\height}
		{\includegraphics[width=\linewidth]{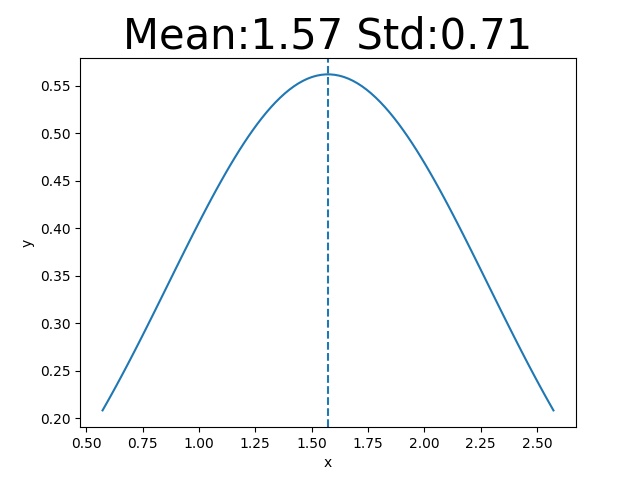}}
    \end{minipage}    
        
    \\  
    
    \hline
    	
    \end{tabular}
    \centering                
    
    \label{tab: ABBS example}               
}
\end{table*}





\begin{algorithm}[h]

    \caption{Maximum zoom-in  factor}
    \label{alg: Maximum zoom in}
    \KwIn{
            $Box1$,
            $Box2$,
            $ScaleMin$,
            $ScaleMax$.
            
    }
    \KwOut{$ScaleMid$, the maximum zoom-in  factor.}
     Init $IterNum$;
     
    \While{$IterNum > 0$}{
        
		$ScaleMid \gets (ScaleMin +  ScaleMax) \div   2 $\;
		                        
		$Box1Scaled  \gets Box1$ zoom-in with scale $ScaleMid$  \;
		
		$Box2Scaled  \gets  Box2$ zoom-in with scale $ScaleMid$  \;
		
		$BoxIou \gets IOU(Box1Scaled, Box2Scaled)$ \;
		
		\If{$BoxIou < IterThreshold$}{
		    return $ScaleMid$
		}

		\eIf{$BoxIou > 0$}
        {
        
           $ ScaleMax \gets ScaleMid $
            
        }{
        
           $ ScaleMin \gets ScaleMid $ 

		}
		
        $IterNum \gets IterNum - 1$ \;
        
	}
\end{algorithm}

\begin{algorithm}[h]
    \caption{Potential relationship detection via ABS}
    \label{alg: ABBSUse}
    \KwIn{$RelationPairs[Cate_A$,$Cate_B$,$Box_A$,$Box_B]$}
    \KwOut{$Scores$}
    initialize $Scores \gets []$ \;
    \For{ $ Cate_A,Cate_B,Box_A,Box_B$ in $RelationPairs$}{
        \uIf{$M(Cate_A,Cate_B)=$None}{
            continue
        }
        
        $IOU_{AB}$ = $ IOU(Box_A,Box_B)$
        
        \uIf{$IOU_{AB}>=0$}{
        
            \uIf{$Box_A$ include $Box_B$}{
              \uIf{$M_c(Cate_A,Cate_B)>0$}{
                $Scores.add (IOU_{AB})$
              }\Else{
                 continue
              }
            }
            
            \uElseIf{$Box_B$ include  $Box_A$}{
              \uIf{$M_c(Cate_B,Cate_A)>0$}{
                $Scores.add (IOU_{AB})$
              }\Else{
                 continue
              }
            }
            
            \Else{
              $P_{down} \gets  MinDown(Box_A, Box_B)$ \;
              \uIf{$P_{down}$   follow  $M_d(Cate_A,Cate_B)$}{
                $Scores.add (IOU_{AB})$
              }\Else{
                 continue
              }
    
            }
        }\Else{
            $P_{up} \gets MaxUp(Box_A, Box_B)$  \;
            \uIf{$P_{up}$ follow $M_u(Cate_A,Cate_B)$}{
                $Scores.add(0)$
            }\Else{
                continue}}}
    $Sort(Scores)$
\end{algorithm}

$M_d(A, B)$ and $M_u(A, B)$ are recorded as the adaptive bounding box scaling factor  of the corresponding bounding boxes of the two objects.  The minimum zoom-out factor and maximum zoom-in  factor are the scale factors of two objects whose bounding boxes intersect exactly when they are enlarged or reduced with the same scale.
Take the calculation of the  maximum zoom-in  factor as an example:  when $Box_A$ and $Box_B$ are the original scale size (where A and B denote two persons), they are currently non-intersecting. However, when the two bounding boxes are zoomed in at the same scale, $Box_A$ and $Box_B$ will intersect exactly. For instance, at a zoom-in factor of $L$  (refer to Fig. \ref{fig abbsT}), $BoxL_A$ and $BoxL_B$ intersect precisely at the red line, while the intersection area is 0.   Here, the maximum zoom-in factor $L$ is achieved, thereby becoming the maximum zoom-in factor of $Cate_A$ and $Cate_B$.

To calculate the maximum zoom-in factor of two non-intersecting objects, Algorithm \ref{alg: Maximum zoom in} is implemented, and the minimum zoom-out factor of two intersecting objects can be calculated similarly. The calculation of the scaling factor necessitates the consideration of numerous special cases due to the complexity of the relative position of two bounding boxes, resulting in elevated algorithm complexity. To reduce the complexity of the algorithm, we draw inspiration from the bisection method and utilize approximations rather than exact scaling coefficients. 






In the dictionary $M$, the priori information is represented by keywords consisting of $(Cate_A, Cate_B)$, as an exemplar.  $M_C(Cate_A, Cate_B)$ and $M_C(Cate_B, Cate_A)$ are converted to frequencies. Meanwhile, $M_d(Cate_A, Cate_B)$ and $M_u(Cate_A, Cate_B)$  are transformed into Gaussian distributions, to calculate the means and variances of $M_d(Cate_A, Cate_B)$ and $M_u(Cate_A, Cate_B)$, respectively.  Table \ref{tab: ABBS example} displays four examples from $M$, and Algorithm \ref{alg: ABBSUse} elaborates the approach to utilizing $M$ for potential relationship detection. Assume that  $n$ objects are detected in an image; there are a total of $n\times(n-1)$ object pairs. Each object pair can be represented by $Cate_A, Cate_B, Box_A, Box_B$, where A and B correspond to the subject and object, respectively. Our objective is to filter out $m$ object pairs from $n\times(n-1)$ pairs, with the most probable relationship between each pair.  If the $M$ does not contain any stored records of category pair  $<Cate_A, Cate_B>$  corresponding to the object pair $(A, B)$, then there is an assumption that there is no relationship between A and B. If  the $M$ does store $<Cate_A, Cate_B>$,  then it is necessary to discuss the position distribution of $Box_A, Box_B$:

\begin{itemize}
\item In the case of "non-intersection" (as depicted Fig. \ref{fig: BBox four kinds} a),  Algorithm \ref{alg: Maximum zoom in} is employed  to calculate the  maximum zoom-in factor $P_{up}$ of $Box_A$ and $Box_B$. If $P_{up}$  adheres to the distribution of $M_u(Cate_A, Cate_B)$, objects A and B are considered to have an existing relationship; otherwise, if $P_{up}$ falls outside the distribution, they are not considered to have a relationship.   To compute the probability density function for calculating the possibility of the value in a given distribution, integration is required. To reduce computational effort, we use an alternative method to discern if $P_{up}$ is less than the maximum value of $M_u(Cate_A, Cate_B)$. Specifically, if $P_{up}$ is lesser than the maximum value, the relationship between A and B is taken into account, if not, no connection exists between A and B.



\item  In the case of "intersection" (as depicted Fig.  \ref{fig: BBox four kinds} b),  Algorithm \ref{alg: Maximum zoom in} can be adapted slightly to calculate the minimum zoom-out factor $P_{down}$ of $Box_A$ and $Box_B$.  If $P_{down}$ conforms to the distribution of $M_d(Cate_A, Cate_B)$, objects A and B are considered to have an existing relationship, otherwise, if $P_{down}$ falls outside the distribution, they are not considered to have a relationship. 


\item  In the case of "containment " (as depicted Fig.  \ref{fig: BBox four kinds} c and d), check  $M_c(Cate_A, Cate_B)$ or $M_c(Cate_B, Cate_A)$ in $M$. If $M_c(Cate_A, Cate_B)$ is net zero, there exists a relationship where A contains B. Similarly, if $M_c(Cate_B, Cate_A)$ is not zero, there exists a relationship where B contains A.

\end{itemize}

Finally, the object pairs that pass the filtering threshold are ranked by their IOU scores. During the testing phase, the order used to evaluate the model is based on the IOU scores rather than the probability of relation classes.

\begin{figure*}[ht]
\centering
\includegraphics[width=0.95\linewidth]{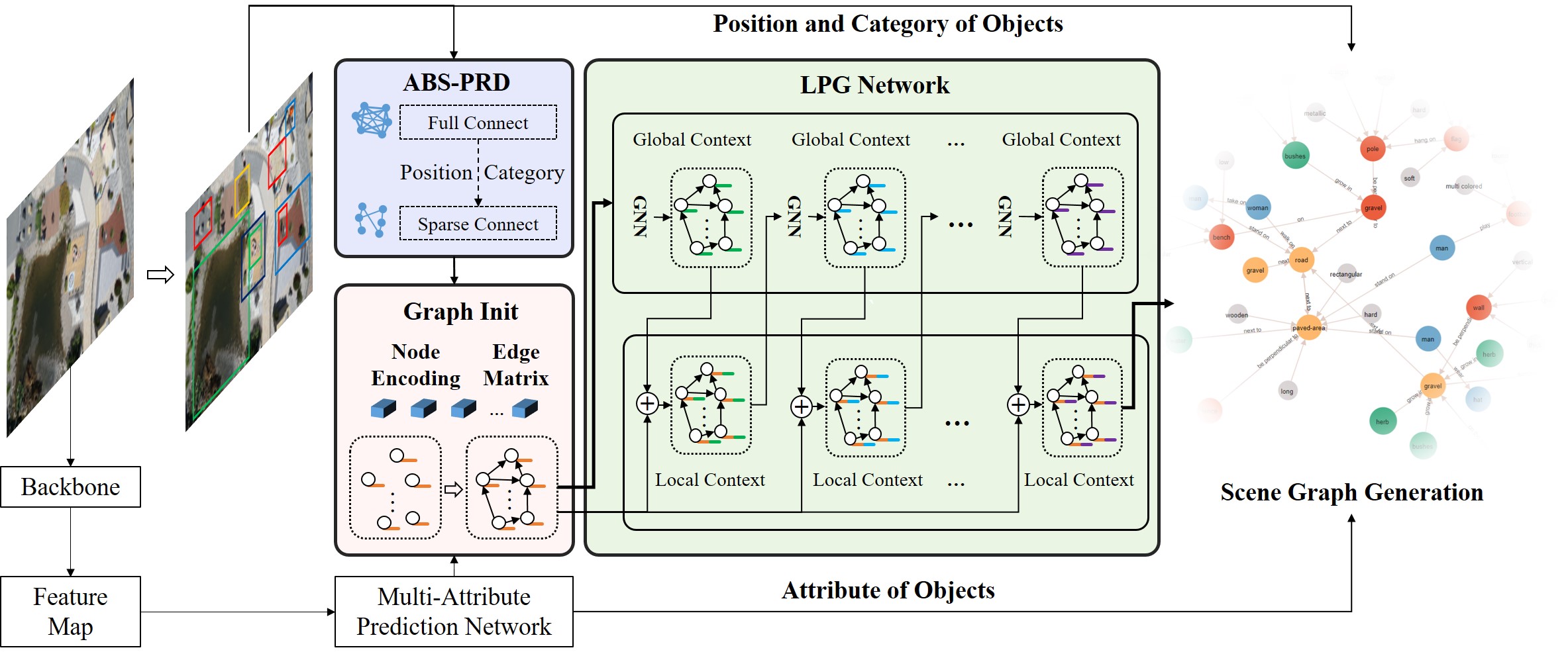}
\caption{The overview of our proposed LPG method. Firstly, a feature map is obtained through object detection of the original image, followed by the utilization of the \textbf{ABS-PRD} module to filter and classify the relevant pairs for sparse computation. Subsequently, the \textbf{Graph Init} module initializes the scene graph. Lastly, through the \textbf{LPG}, the model captures local information and high-order semantic information of the neighborhood simultaneously, generating precise feature vectors of the relational categories to the classifier.}

\label{fig: model}
\end{figure*}

Compared to IOU-PRD, ABS-PRD offers advantages in detecting and filtering object pairs. It effectively identifies non-intersecting yet related object pairs and filters out a portion of intersecting but non-related object pairs. This leads to improved accuracy in determining the true relationships between objects. ABS-PRD significantly enhances the model's performance in detecting potentially related object pairs, especially when dealing with a large number of objects in the images with the overhead view. It can serve as a replacement for relationship detection methods in various SGG techniques, such as IOU-PRD. As a general statistical approach, ABS-PRD provides object pairs that are more likely to be related, thereby improving the performance of the SGG model.

\subsection{Locality-preserving Graph Convolutional  Network}

Our proposed method LPG for the AUG task utilizes information from two subtasks: object detection and multi-label attribute prediction. For more accurate prediction of objects at different scales, we use Cascade R-CNN instead of Faster-RCNN in the object detection task. Alternatively, a more efficient algorithm during the object detection phase could also be used. The multi-label attribute prediction task uses a feedforward neural network. Semantic attribute information from multi-labeled attributes is introduced as an intermediate quantity between the low-level features of the original data and the high-level prediction results, improving the accuracy of relational category prediction and the interpretability of the model. The LPG is designed with an integrated structure that can effectively combine the local context and neighborhood information at the information aggregation step. The following is an example of the locality-preserving network ( note $l$ as the number of layers).

After object detection and multi-label attribute prediction, the attribute $\vec{h}_{att}$, position $\vec{h}_{box}$ and category information $\vec{h}_{cls}$ of the nodes are obtained. We concatenate these features based on their dimensions to create the feature vectors for each object.
{
\setlength{\abovedisplayskip}{0.3cm}
\setlength{\belowdisplayskip}{0.3cm}
\begin{align}
\centering
\vec{h} = [\vec{h}_{att},\vec{h}_{box},\vec{h}_{cls}]
\label{alg:4.1}
\end{align}}

\noindent 
where  $[\cdot]$ denotes the concatenate operation. Thus the objects can be represented as $   \mathcal{V}_{local} =\left \{ \vec{h}_{1} ,\vec{h}_{2} ,...,\vec{h}_{n}  \right \},\vec{h}_{i}\in \mathbb{R}^{F}$,  where $n$ denotes the number of nodes, $F$ denotes the dimension of the node  feature vector. 

Then all possible object pairs $n\times(n-1)$, including their respective object categories and bounding boxes, are enumerated from the $n$ objects.  These object pairs are passed through ABS-PRD to identify pairs that have possible relations. Directed edges $  \mathcal{E}  =\left \{ \vec{h}_{i} \to \vec{h}_{j}  \right \},  \vec{h}_{i}  \in \mathbb{R}^{F},  \vec{h}_{j}  \in \mathbb{R}^{F}$ are created for each object pair, which contains subject and object orders. Therefore, each image is initialized with a sparse graph structure $\mathcal{G} =\left (\mathcal{V}_{local},\mathcal{E} \right )$.  The feature matrix $H^{0}$ of the nodes can be seen as   $ \mathcal{V}_{local}$, and we encode $\mathcal{E}$ as the adjacency matrix $A$ for the graph.


The LPG utilizes a multi-layer graph neural network approach for message passing and aggregation. Each layer in the network contains a message passing and aggregation operation, as well as an integration operation,  which can be expressed as:
{
\setlength{\abovedisplayskip}{0.3cm}
\setlength{\belowdisplayskip}{0.3cm}
\begin{align}
\centering
    H^{l+1} = \sigma (AH^{l}W)
\label{alg:4.4}
\end{align}
}

\noindent where $\sigma$ is the nonlinear activation function, $W \in \mathbb{R}^{F}$ is a weight matrix with random initialization. After the graph convolutional  operation on $\mathcal{G}$, the feature vector of nodes can be expressed as  $\mathcal{V}_{global} = \left \{ \vec{h}_{1}^{'},\vec{h}_{2}^{'},...,\vec{h}_{n}^{'}\right \},\vec{h}_{i}^{'}\in \mathbb{R}^{F^{'}}$, where $F^{'}$ denotes the size of the feature vector computed by the layer of the graph convolutional  network, and the last step is to concatenate $\mathcal{V}_{local}$ and $\mathcal{V}_{global}$ into $\mathcal{V}_{c}$ by feature  dimension, where $ [\vec{h}_{i},\vec{h}_{i}^{'}] \in \mathbb{R}^{F+F^{'}}$.
{
\setlength{\abovedisplayskip}{0.3cm}
\setlength{\belowdisplayskip}{0.3cm}
\begin{align}
\centering
    \mathcal{V}_{c} = \left \{ [\vec{h}_{1} ,\vec{h}_{1}^{'}] ,[\vec{h}_{2} ,\vec{h}_{2}^{'}] ,...,[\vec{h}_{n} ,\vec{h}_{n}^{'}] \right \} 
\label{alg:4.6}
\end{align}
}

The previous equations and steps are the complete calculation for one layer of the LPG, and the global context information is gradually obtained through a stacked structure of multiple LPG layers. The output of the message passing through the previous layer is used as the input to the next layer. It is important to note that $[ \mathcal{V}_{local}$,$\mathcal{V}_{golbal} ]$ is the input to the object feature matrix in the non-first layer, while in the first layer, only $\mathcal{V}_{local}$ is used.  After multiple message passing layers, the output vector from the final layer $\mathcal{V}_{global}^{'}$ is concatenated with $\mathcal{V}_{local}$ as the feature vector for relationship prediction, denoted as $\mathcal{V}_{c}\in \mathbb{R}^{F+{F}^{'}}$. 

The ABS-PRD filters object pairs with potential relationships selected from $\mathcal{V}_{c}$ among $n\times(n-1)$ object pairs involving $n$ nodes that could have relationships. The pairs of objects with potential relationships $\mathcal{P}$ are represented as:
{
\setlength{\abovedisplayskip}{0.3cm}
\setlength{\belowdisplayskip}{0.3cm}
\begin{align}
\centering
    \mathcal{P}= \left \{ [ \mathcal{V}_{c_i} ,\mathcal{V}_{c_j}] \right \}  \in \mathbb{R}^{k\times2(F+{F}^{'})}
\label{alg:4.7}
\end{align}
}

\noindent where $\mathcal{V}_{c_i}$ and $\mathcal{V}_{c_j}$ represent feature vectors of nodes $i$ and $j$. $\mathcal{V}_{c_i}\in \mathcal{V}_{c} , \mathcal{V}_{c_j}\in \mathcal{V}_{c} $ , $i\in [1,n], j \in [1,n]$ and $k \in [0,n\times (n-1)]$ is number of $\mathcal{P}$. 

Using the filtered and potential object pairs $\mathcal{P}$ obtained from $\mathcal{V}_{c}$, we feed the corresponding comprehensive features into the fully connected layer to predict the predicate. The predicate classifier follows this structure:
{
\setlength{\abovedisplayskip}{0.3cm}
\setlength{\belowdisplayskip}{0.3cm}
\begin{align}
\centering
   R_{pred} = \mathrm{Softmax} \left ( \mathrm{FFN} (\mathcal{P}) \right ) 
\label{alg:4.8}
\end{align}
}

\noindent where $\mathrm{FFN}$ represents the fully connected layer,  after activation by the $\mathrm{Softmax}$ function, the prediction result $R_{pred}$ is obtained. Therefore, the final loss $ \mathcal{L}$ of the LPG can be expressed as:
{
\setlength{\abovedisplayskip}{0.3cm}
\setlength{\belowdisplayskip}{0.3cm}
\begin{align}
\centering
   \mathcal{L}  = \mathrm{CrossEntropy}( R_{pred}, R_{gt} )
\label{alg:4.9}
\end{align}
}

For each object pair $\mathcal{P}$, the first object $\mathcal{V}_{c_i}$ is identified as the $\left \langle subject \right \rangle$, while the second object $\mathcal{V}_{c_j}$ is identified as the $\left \langle object \right \rangle$. The relationship predicate that corresponds to the $\mathcal{P}$ is noted as $R_{gt}$. The final loss $\mathcal{L}$ of the LPG is calculated using the cross-entropy loss function $\mathrm{CrossEntropy}$.

\section{Experimental analysis and discussions}
\label{sec5}

This section presents experiments related to the scene graph generation task, showing the performance results of the LPG and five mainstream methods on the AUG dataset. Also, experiments on ablation studies are conducted and used to verify the effectiveness of important modules of our method. This section also includes visual results (shown in Fig. \ref{img:result}) for better understanding and demonstration.

\subsection{Experiment Settings}

The experiments related to the scene graph generation task are performed using the AUG dataset. We set the main hyperparameters of all the relevant experiments as follows:

We use the stochastic gradient descent 
optimizer to optimize all tasks, and the Cascade R-CNN object detection model algorithm is implemented using the MMDetection \cite{MMDet} framework. The parameters for object detection in our experiments are as follows: batch size 4, learning rate 0.003, and mean and variance statistics in image pre-processing are adapted to the dataset itself. Other parameters such as anchor point generation, post-processing, and prediction IOU thresholds are also adapted depending on the dataset.

The SGCls task uses the ImageNet pre-training Resnet50 \cite{Resnet} model parameters for model initialization for the object classification task. The batch size is 6 with a learning rate of 0.01 and a uniform input image size of 224. For the three SGG subtasks, the parameters are set as follows: the batch size of 2, the learning rate of 5, and an IOU threshold of 0.5 for the effective prediction of object positions in the SGDet task evaluation required and the true object positions. Following the work of Motif Net \cite{motif}, mR@K and R@K are chosen as evaluation metrics, where K is set to {50, 150, 250}, and 50 is set because it is close to the average number of objects in AUG images. For the attribute prediction multi-label classification task, batch size 64, learning rate 0.01, and optimizer configurations similar to the above are chosen. The experiments are run on an Intel Xeon Silver 4210 CPU, TITAN RTX GPU, and a seed of 2022 is used.

\begin{table*}[ht]
\setlength{\tabcolsep}{15pt}
\renewcommand\arraystretch{1.2}
\caption{The performances of different SGG methods in the AUG dataset on R@K(\%).}
\centering
\resizebox{\textwidth}{!}{%
\begin{tabular}{c|ccc|ccc|ccc}
\hline
\multirow{2}{*}{Models} & \multicolumn{3}{c|}{PredCls} & \multicolumn{3}{c|}{SGCls} & \multicolumn{3}{c}{SGDet} \\ \cmidrule{2-10} 
 & 50 & 150 & 250 & 50 & 150 & 250 & 50 & 150 & 250 \\ \hline
Motif & 26.4 & 35.8 & 38.9 & 20.2 & 25 & 26 & 12.8 & 16.2 & 17.4 \\
VCTree & 29.1 & 37.7 & 40.5 & 22 & 26.6 & 28.6 & 11.5 & 13.4 & 15.1 \\
GPSNet & 28.2 & 37.2 & 39.2 & 17.2 & 22.5 & 24.3 & 13.1 & 17.1 & 18.8 \\ 
TDE (Motif) & 33 & 43.2 & 46.3 & 20.6 & 25.2 & 26.5 & 12.7 & 15.5 & 16.9 \\
TDE (VCTree) &  33.7 & 44.7 & 48.4 & 22.8 & 26.7 & 28.9 & 13.9 & 17.6 & 18.3 \\
EBM (Motif) & 31 & 41.7 & 45.3 & 18.3 & 23.1 & 25.6 & 11.6 & 14.3 & 15 \\
DLFE (Motif) & 25.9 & 36.2 & 39.9 & 15.9 & 21.1 & 22.8 & 11.8 & 14.1 & 16 \\
IETrans (Motif) & 22.9 & 30.7 & 33.7 &  2.8 & 3.4 & 3.7 & 0.7 & 1.4 & 1.5 \\ \hline
\textbf{Our LPG with Faster RCNN} & \textbf{35.3} & \textbf{49.1} & \textbf{52.3} & \textbf{19.8} & \textbf{28.0} & \textbf{30.0} & \textbf{14.0} & \textbf{22.1} & \textbf{24.7} \\
\textbf{Our LPG with Cascade RCNN} & \textbf{40.1} & \textbf{55.0} & \textbf{57.8} & \textbf{24.9} & \textbf{33.6} & \textbf{35.3} & \textbf{16.3} & \textbf{24.4} & \textbf{27.0} \\\hline
\end{tabular}%

\label{table:overall1}}
\end{table*}

\begin{table*}[ht]
\setlength{\tabcolsep}{15pt}
\renewcommand\arraystretch{1.2}
\caption{The performances of different SGG methods in the AUG dataset on mR@K(\%). }
\centering
\label{tab:my-table}
\resizebox{\textwidth}{!}{%
\begin{tabular}{c|ccc|ccc|ccc}
\hline
\multirow{2}{*}{Models} & \multicolumn{3}{c|}{PredCls} & \multicolumn{3}{c|}{SGCls} & \multicolumn{3}{c}{SGDet} \\ \cmidrule{2-10} 
 & 50 & 150 & 250 & 50 & 150 & 250 & 50 & 150 & 250 \\ \hline
Motif & 11.3 & 15.8 & 18.1 & 5.1 & 6.9 & 7.2 & 3.8 & 4.7 & 5.3 \\
VCTree & 12.5 & 15.8 & 17.6 & 5.2 & 7.6 & 8.6 & 3.1 & 3.7 & 4.8 \\
GPSNet & 12.0 & 16.5 & 17.6 &  6.5 & 9.1 & 10.1 & 4.6 & 5.4 & 6.2 \\
TDE (Motif) & 11 & 14.6 & 16.6 & 5.2 & 6.3 & 6.7 & 3 & 4.5 & 5.4 \\
TDE (VCTree) & 11.3  & 18.1 & 19.5 & 4.8 & 6.2 & 6.7 & 3.3 & 4.4 & 4.8 \\
EBM (Motif) & 10.5 & 14.6 & 16.5 & 3.2 & 5.1 & 5.9 & 2.1 & 2.7 & 2.8 \\
DLFE (Motif) & 9.1 & 15.1 & 17 & 3.2 & 4.8 & 5.3 & 4 & 4.4 & 4.9 \\
IETrans (Motif) & 11.4 & 14.5 & 15.6 & 1.1 & 1.4 & 1.4 & 0.2 & 0.3 & 0.3 \\ \hline
\textbf{Our LPG with Faster RCNN} & \textbf{14.2} & \textbf{19.4} & \textbf{22.1} & \textbf{7.1} & \textbf{9.8} & \textbf{10.9} & \textbf{4.6} & \textbf{6.5} & \textbf{7.7} \\
\textbf{Our LPG with Cascade RCNN} & \textbf{22.4} & \textbf{29.5} & \textbf{31.1} & \textbf{14.5} & \textbf{17.2} & \textbf{18.3} & \textbf{7.1} & \textbf{9.5} & \textbf{10.9} \\ \hline
\end{tabular}%
}
\label{table:overall2}
\end{table*}

\subsection{Sensitivity Analysis of the Hyperparameter}
\label{sec:5.2}

The number of layers in the LPG is a hyperparameter. Too few layers can lead to the inadequate acquisition of neighborhood information, while an overabundance of layers can result in less differentiated node information. As such, it is necessary to determine the optimal number of layers in the LPG.

\begin{table}[htpb]
\setlength{\tabcolsep}{5pt}
\renewcommand\arraystretch{1.2}
\caption{The hyperparameter  analysis in the AUG dataset on R@K(\%).}
\centering
\resizebox{\columnwidth}{!}{%
\begin{tabular}{l|ccc|ccc|ccc}
\hline
 & \multicolumn{3}{c|}{PredCls} & \multicolumn{3}{c|}{SGCls} & \multicolumn{3}{c}{SGDet} \\ \cmidrule{2-10} 
 & 50 & 150 & 250 & 50 & 150 & 250 & 50 & 150 & 250 \\ \hline
$l$=1 & 29 & 42 & 44.5 & 19.1 & 28.2 & 30.5 & 9.2 & 16.2 & 18.1 \\
$l$=\textbf{2} & \textbf{33.1} & \textbf{47.1} & \textbf{49.7} & \textbf{21.1} & \textbf{31.8} & \textbf{33.7} & \textbf{14.5} & \textbf{23.7} & \textbf{26.2} \\
$l$=3 & 29.9 & 44.5 & 47.2 & 16.5 & 23.9 & 25.5 & 12.6 & 19.5 & 22
\\ \hline
\end{tabular}%
\label{tab:5.2.1}
}
\end{table}
\begin{table}[htpb]
\setlength{\tabcolsep}{5pt}
\renewcommand\arraystretch{1.2}
\caption{The hyperparameter analysis in the AUG dataset on mR@K(\%).}
\centering
\resizebox{\columnwidth}{!}{%
\begin{tabular}{l|ccc|ccc|ccc}
\hline
 & \multicolumn{3}{c|}{PredCls} & \multicolumn{3}{c|}{SGCls} & \multicolumn{3}{c}{SGDet} \\ \cmidrule{2-10} 
 & 50 & 150 & 250 & 50 & 150 & 250 & 50 & 150 & 250 \\ \hline
$l$=1 & 12 & 15.8 & 16.6 & 7.6 & 11.3 & 12.1 & 3 & 4.7 & 5.3 \\
$l$=\textbf{2} & \textbf{18.3} & \textbf{24.8} & \textbf{26.3} & \textbf{11.5} & \textbf{16.4} & \textbf{17.4} & \textbf{6.6} & \textbf{8.9} & \textbf{9.5} \\
$l$=3 & 9.9 & 13.9 & 15.3 & 9.2 & 12.2 & 13.8 & 3.9 & 5.7 & 7.1 \\
\hline
\end{tabular}%
\label{tab:5.2.2}
}
\end{table}

Based on the current GNN models, the  number of GNN layers  is generally between 1 and 4. The layer number setting is related to the neighborhood density of the nodes. 
The evaluation results of the LPG on the three subtasks of the scene graph generation method are presented by adjusting the layer hyperparameters (1, 2, and 3 layers respectively, as seen in Table \ref{tab:5.2.1} and Table \ref{tab:5.2.2}). The experimental results indicate that the most effective performance is achieved with 2 layers of the locality-preserving graph convolutional network.

\subsection{Comparison with the State-of-the-art Methods}

\begin{figure*}[ht!]
\centering
\includegraphics[width=0.98\linewidth]{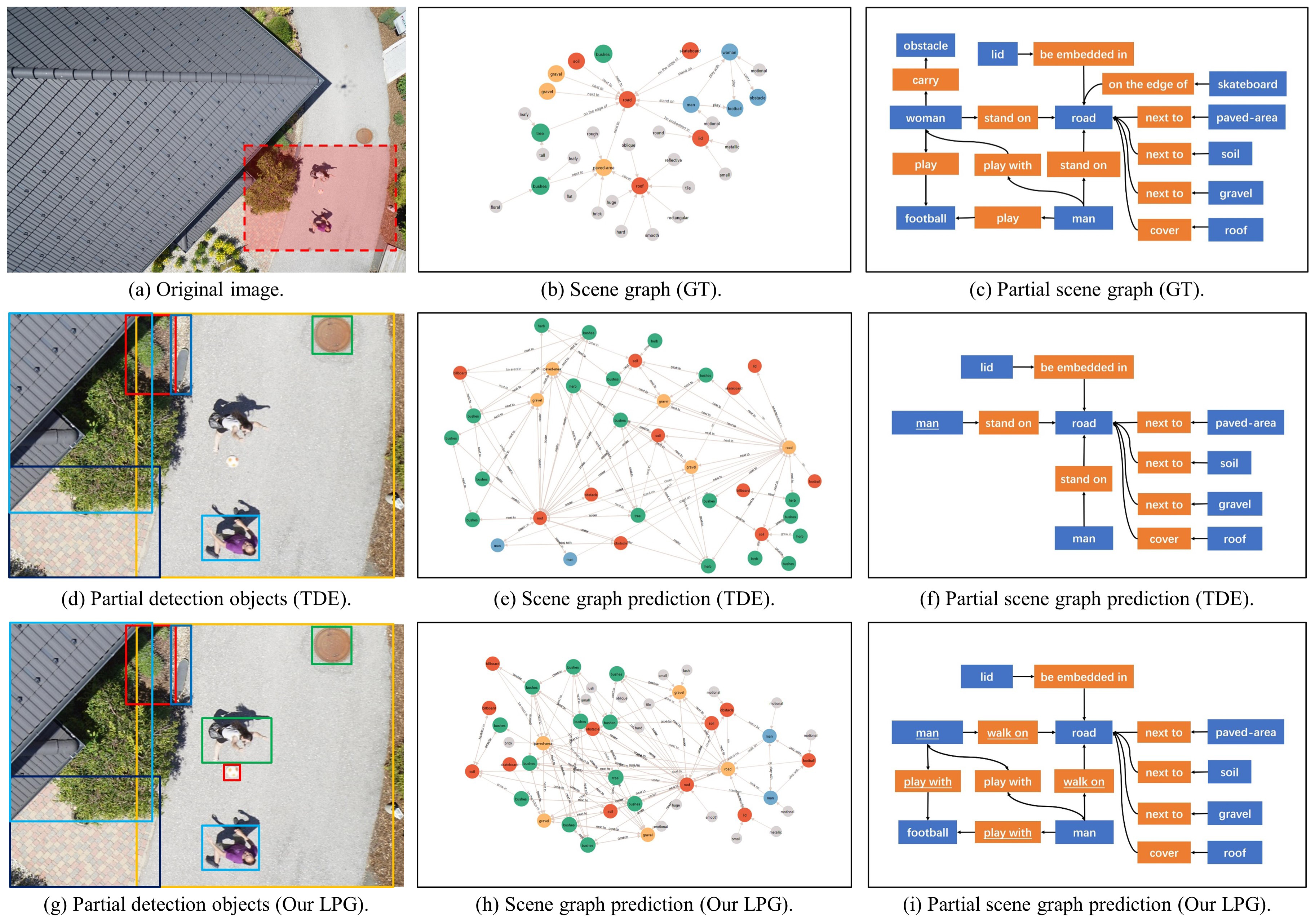}
		
\caption{This is an example from the SGDet task on the test dataset, where GT refers to ground truth: (a) shows the original image, (b) shows the labeling information, presented as a scene graph, (c)  shows the partial scene graph corresponding to real label information, (d) shows the partial image with object detections using TDE with VCTree, (e) shows the prediction information of scene graph using TDE with VCTree, (f) shows the  partial prediction scene graph using TDE with VCTree, (g) shows the partial image with object detections using our method, (h) shows the prediction information of scene graph using our method, (i) shows the  partial prediction scene graph using our method.}

\label{img:result}
\end{figure*}

In Section \ref{sec:5.2}, we set the hyperparameters for the experiments. Based on these settings, we present and analyze the experimental results below. Two tables containing the results for the LPG model and mainstream scene graph generation models on the AUG dataset with regard to two metrics are presented. In the following, Model A (Model B) refers to the optimization approach using Model A with Model B acting as the backbone network.


To verify the validity of our method, we compare it to several current state-of-the-art methods, which include competitive backbone networks and post-processing methods. The independent backbone network methods we choose to compare against are Motif \cite{motif}, VCTree \cite{vctree}, and GPSNet \cite{GPSNet}. TDE \cite{TDE}, DLFE \cite{DLFE}, EBM \cite{EBM}, and IETrans \cite{IETrans} are all methods for post-processing data and inference based on the backbone network. Among these, TDE is a novel SGG framework based on causal inference rather than conventional likelihood, and DLFE helps solve the long-tailed bias issue by balancing high-frequency and low-frequency prediction accuracy using prior statistical information to correct category probability distribution. IETrans aims to alleviate the data distribution problem by automatically creating an enhanced dataset with more adequate and consistent annotations for all predicates.

Overall, our LPG method proposed in this study achieves the best results in the R@K and mR@K metrics for the PredCls, SGCls, and SGDet tasks, outperforming all other methods (as shown in Table \ref{table:overall1} and Table \ref{table:overall2}). Despite outperforming baseline models, post-processing methods like IETrans and DLFE perform poorly on the AUG dataset due to over-fitting and data quality issues. Backbone networks such as VCTree and GPSNet have shown consistently excellent performance on both VG and AUG datasets. Lastly, we note that TDE, a data processing method used during inference, improves the performance of baseline models on datasets with different data distributions without directly affecting inference results through prior dataset information.


The results of the scene graph generation are visualized in Fig. \ref{img:result}.  Comparing the scene graph of the real image and the scene graph of the prediction result, we find that the triad information in the prediction result is greater than that in the actual annotation.  It is worth noting that, in the actual prediction result, the sparseness of the real annotation implies that a correct pair of object relationships may not improve the metric. Additionally, two individuals are playing football inside a red-boxed region in Fig. \ref{img:result} \textbf{a}. Although the bounding boxes of the objects within the boxed region do not intersect with each other in pairs, the ABS-PRD successfully detects all three objects (see \textbf{i}); however, other methods using IOU-PRD cannot do so intrinsically (see \textbf{f}).

\subsection{Ablation Study}

Three groups of ablation experiments are required to verify the validity of the LPG method proposed in this paper. Table \ref{tab:ablation1} and Table \ref{tab:ablation2} show the results of these experiments. 'LPG w/o Attr' refers to a method that does not utilize attribute information, while 'LPG w/o ABS' uses the IOU-PRD instead of the ABS-PRD in the stage of potential relationship detection. 'LPG w/o LP' indicates a strategy that does not make the local context preservation.


\textbf{Attribute Embedding.} The model's prediction performance after the introduction of attributes improved on average by 55\%, 42\%, and 86\% on R@K, respectively, and by 225\%, 130\%, and 196\% on mR@K, respectively. In conclusion, the addition of attributes has greatly improved the model performance of the scene graph generation task, with the most significant improvement in the SGDet task. This may be due to the fact that among the three tasks, the higher the model performance on the PredCls and SGCls tasks, the more it relies on accurate relation detection. In these two tasks, attribute information can only play an auxiliary role as part of the features embedded in the initial nodes. However, in the SGDet task, the model cannot obtain the real object category and object position information. The prediction results of object attributes can be used as an intermediate quantity for high-dimensional object semantics, thus becoming the main role in improving performance.

\begin{table}[t!]
\setlength{\tabcolsep}{3pt}
\renewcommand\arraystretch{1.3}
\caption{The ablation study in the AUG dataset on R@K(\%).}
\centering
\resizebox{\columnwidth}{!}{%
\begin{tabular}{c|ccc|ccc|ccc}
\hline
 \multirow{2}{*}{Models} &  & PredCls &  &  & SGCls &  &  & SGDet &  \\ \cmidrule{2-10} 
 & 50 & 150 & 250 & 50 & 150 & 250 & 50 & 150 & 250 \\ \hline
LPG w/o Attr & 24.2 & 36.7 & 38.5 & 16.1 & 24.5 & 25.9 & 8.5 & 12.8 & 15.3 \\
LPG w/o ABS & 33.1 & 47.1 & 49.7 & 21.1 & 31.8 & 33.7 & 14.5 & 23.7 & 26.2 \\
LPG w/o LP  & 15.9 & 21.2 & 22.1 & 9.38 & 13.8 & 14.5 & 7.3 & 12.6 & 14.2 \\
\textbf{Our LPG} & \textbf{40.1} & \textbf{55.0} & \textbf{57.8 }& \textbf{24.9} & \textbf{33.6} & \textbf{35.3} & \textbf{16.3} & \textbf{24.4} & \textbf{27.0} \\ \hline
\end{tabular}
\label{tab:ablation1}
}
\end{table}

\begin{table}[t!]
\setlength{\tabcolsep}{3pt}
\centering
\caption{The ablation study in the AUG dataset on mR@K(\%).}\label{tab:ablation2}
\renewcommand\arraystretch{1.3}
\resizebox{\columnwidth}{!}{%
\begin{tabular}{c|ccc|ccc|ccc}
\hline
\multirow{2}{*}{Models} &  & PredCls &  &  & SGCls &  &  & SGDet &  \\ \cmidrule{2-10}
& 50 & 150 & 250 & 50 & 150 & 250 & 50 & 150 & 250 \\ \hline
LPG w/o Attr & 6.3 & 9.4 & 10.1 & 5.5 & 8 & 8.6 & 2.2 & 3.4 & 3.8 \\
LPG w/o  ABS & 18.3 & 24.8 & 26.3 & 11.5 & 16.4 & 17.4 & 6.6 & 8.9 & 9.5 \\
LPG w/o LP & 7.0 & 8.3 & 8.7 & 2.4 & 3.1 & 3.7 & 1.7 & 3.1 & 3.5 \\
\textbf{Our LPG} & \textbf{22.4} & \textbf{29.5} & \textbf{31.1} & \textbf{14.5} & \textbf{17.2} & \textbf{18.3} & \textbf{7.1} & \textbf{9.5} & \textbf{10.9} \\ \hline
\end{tabular}
}
\end{table}

\textbf{ABS-PRD.} Compared with the IOU-PRD, the ABS-PRD can provide more pairs of candidate objects containing positive samples. Specifically, ABS-PRD can effectively detect relationships among non-intersecting objects, even over long distances. It can also eliminate intersecting but meaningless object pairs. The experimental results demonstrate that the ABS-PRD outperforms the IOU-PRD in all three sub-tasks. ABS-PRD achieves an average improvement of 6.47\% on R@K and 23.3\% on mR@K.

\textbf{Local context Preserving.} The node initialization strategy adopted in this paper contains practical implications, including multi-label attributes, categories, and locations of objects. The LPG leverages a locality-preserving structure to fully preserve the completeness of the initial node information. The use of a locality-preserving structure ensures that the local features of the node are fully preserved while aggregating neighborhood information, which achieves the most significant effect enhancement, an average improvement of 136.9\% on R@K and 313.5\% on mR@K.


\section{Conclusion}
\label{sec6}

This paper constructs and releases the AUG dataset including object, attribute, and relationship annotations, which is the first AUG dataset to the best of our knowledge. To avoid the local context being overwhelmed in the complex aerial urban scene, we propose the LPG, a targeted method based on our AUG dataset and similar data. The LPG combines the non-destructive initial features of objects with the scene information to predict object relationships. To address the problem that there exists an extra-large number of potential object relationship pairs but only a small part of them is meaningful, we propose the ABS-PRD instead of the IOU-PRD to effectively improve the detection accuracy of the PRD stage.  Extensive overall experiments on the AUG dataset show that the LPG can significantly outperform 8 state-of-the-art methods. We hope these experiments can act as benchmarks for fair comparisons between the AUG methods. In addition, extensive ablation experiments verify the effectiveness of each of the key modules of the LPG, providing a reference for future AUG approaches. We anticipate that the AUG dataset and proposed LPG will advance the research and development of the AUG task.



\section*{Acknowledgment}
This work was supported by the National Natural Science Foundation of China under Grants 41971284 and 42030102; the Special Fund of Hubei Luojia Laboratory under Grant 220100032.

{\small
\bibliographystyle{IEEEtran}
\bibliography{refs}
}

\vfill


\end{document}